\documentclass[a4paper,11pt]{article}
\usepackage[numbers]{natbib}

\usepackage[utf8]{inputenc} 
\usepackage[T1]{fontenc}    
\usepackage{hyperref}       
\usepackage{url}            
\usepackage{booktabs}       
\usepackage{amsfonts}       
\usepackage{nicefrac}       
\usepackage{microtype}      

\usepackage{xcolor}

\usepackage{authblk}        

\DeclareUnicodeCharacter{00A0}{ }

\usepackage{color}
\usepackage{caption}

\usepackage{graphicx}
\usepackage{times}
\usepackage{fullpage}

\usepackage{dsfont}
\usepackage{amsmath,amsfonts,amssymb,amsthm}
\usepackage{mathtools}

\usepackage{tikz}           
\usetikzlibrary{arrows}

\usepackage{array,multirow}

\usepackage{amsmath,amsfonts,bm}

\def\1{\bm{1}}

\DeclareMathAlphabet{\mathsfit}{\encodingdefault}{\sfdefault}{m}{sl}
\SetMathAlphabet{\mathsfit}{bold}{\encodingdefault}{\sfdefault}{bx}{n}

\newcommand{\R}{\mathbb{R}}
\newcommand{\N}{\mathbb{N}}

\newcommand{\softmax}{\mathrm{softmax}}

\DeclareUnicodeCharacter{00A0}{ }

\makeatletter
\def\blfootnote{\xdef\@thefnmark{}\@footnotetext}
\makeatother

\usepackage{scrextend}

\definecolor{myred}{rgb}{0.77, 0.0, 0.1}
\definecolor{newgreen}{RGB}{0,153,0}
\definecolor{myturq}{rgb}{0.1, 0.7, 0.7}
\hypersetup{
	colorlinks,
	urlcolor=blue,
	linkcolor=blue, 
	citecolor=blue, 
	linktoc=all 
}

\title{ZiMM: a deep learning model for long term and blurry relapses with non-clinical claims data}

\author[1]{Anastasiia Kabeshova}
\author[2]{Yiyang Yu}
\author[3]{Bertrand Lukacs}
\author[4]{Emmanuel Bacry}
\author[2,5]{Stéphane Gaïffas}
\affil[1]{CMAP, École polytechnique, Palaiseau, France}
\affil[2]{LPSM, Université de Paris, France}
\affil[3]{Service d'Urologie - Hopital Tenon - APHP}
\affil[4]{CEREMADE, Université Paris Dauphine, PSL, Paris, France}
\affil[5]{DMA, Ecole normale supérieure, Paris, France}

\begin{document}
	
	\blfootnote{Equal contributions from A. Kabeshova and Y. Yu}
	
	\maketitle
	
	\begin{abstract}
		This paper considers the problems of modeling and predicting a long-term and ``blurry'' relapse that occurs after a medical act, such as a surgery.
		We do not consider a short-term complication related to the act itself, but a long-term relapse that clinicians cannot explain easily, since it depends on unknown sets or sequences of past events that occurred before the act.
		The relapse is observed only indirectly, in a ``blurry'' fashion, through longitudinal prescriptions of drugs over a long period of time after the medical act.
		We introduce a new model, called ZiMM (Zero-inflated Mixture of Multinomial distributions) in order to capture long-term and blurry relapses.
		On top of it, we build an end-to-end deep-learning architecture called ZiMM Encoder-Decoder (ZiMM ED) that can learn from the complex, irregular, highly heterogeneous and sparse patterns of health events that are observed through a claims-only database.
		ZiMM ED is applied on a ``non-clinical'' claims database, that contains only timestamped reimbursement codes for drug purchases, medical procedures and hospital diagnoses, the only available clinical feature being the age of the patient.
		This setting is more challenging than a setting where bedside clinical signals are available.
		Our motivation for using such a non-clinical claims database is its exhaustivity population-wise, compared to clinical electronic health records coming from a single or a small set of hospitals.
		Indeed, we consider a dataset containing the claims of almost \emph{all French citizens} who had surgery for prostatic problems, with a history between 1.5 and 5 years.
		We consider a long-term (18 months) relapse (urination problems still occur despite surgery), which is blurry since it is observed only through the reimbursement of a specific set of drugs for urination problems. 
		Our experiments show that ZiMM ED improves several baselines, including non-deep learning and deep-learning approaches, and that it allows working on such a dataset with minimal preprocessing work.
	\end{abstract}

	\section{Introduction}
	\label{sec:introduction}
	
	The increasing volume of Electronic Health Records (EHR) systems of national medical organizations in recent years allows to capture data from millions of individuals over many years. Each individual’s EHR can link data from many sources and hence contain “concepts” such as diagnoses, interventions, lab tests, clinical narratives, and more. This provides great opportunities for data scientists to collaborate on different aspects of healthcare research by applying advanced analytics to these EHR clinical data~\cite{Rajkomar2018}.
	There are several challenges in processing EHR data~\cite{Cheng2016}: data quality, high-dimensionality, temporality which refers to the sequential nature of clinical events, sparsity in both medical codes representation and in timestamp representation, irregularly timed observations, biases such as systematic errors in data collection, and mixed data types with missing data. 
	Representation learning can overcome these challenges~\cite{Rajkomar2018} and the choice of data representation or feature representation plays a significant role in the success of this approach.

	\subsection{Related works}
	\label{sub: related_works}
	
	Recently, much work has been done on developing compact and functional representations of medical records, including the use of deep learning over EHR data~\cite{Bajor2018, Shickel2018}. 
	A notable attempt is stacked denoising autoencoders~\cite{Miotto2016}.
	Denoising autoencoders are also used in~\cite{Beaulieu-Jones2016} to develop patient representation from various binary clinical descriptors on synthetic data.
	Sums of word-level skip-gram embedded vectors of clinical codes are used in~\cite{ChoiY2016} to create full-record representations. Word-level semantic embeddings for diagnosis and intervention codes are constructed in~\cite{Pham2017}, using pooling and concatenation to aggregate embeddings into a vector representing a single admission, while~\cite{Nguyen2016} uses word-level embedding as preprocessing for a CNN (Convolutional Neural Networks) architecture.
	
	Only a few studies were made based on claims data due to its complexity and rare availability.
	Choi et al. learned distributed representations of medical codes (e.g. diagnoses, medications, procedures) from electronic health records (EHRs) and claims data using Skip-gram and applied them to predict future clinical codes and risk groups~\cite{Choi2016}. 
	Cui2vec is a recent study in learning clinical concept embeddings~\cite{Beam2018}, which applied word2vec~\cite{mikolov2013distributed} and Glove~\cite{pennington2014} on multiple medical resources such as structured claims data, biomedical journal articles and unstructured clinical notes. Xiang et al.~\cite{Xiao2018} proposed learning a distributional representation of clinical concepts considering temporal dependencies along the longitudinal sequence of a patient’s records based on claims data.
	
	Other efforts have aimed at encoding temporal aspects of EHR data for predictive tasks. 
	In~\cite{Choi2015}, time-stamped events are used as inputs to a particular type of RNNs (Recurrent Neural Networks) to predict future disease diagnosis, while a graph-based attention model is used in~\cite{Choi2019} to learn concept representations. 
	Another interesting recent effort is~\cite{Rajkomar2018}, which maps raw EHR data to the FHIR format (Fast Healthcare Interoperability Resources~\cite{Bender:2013}) to encode EHR information for several different sequence-oriented models.
	
	RETAIN~\cite{Choi2016retain} and GRAM~\cite{Choi2016gram} are two state-of-the-art models using RNNs for predicting future diagnoses. 
	However, they cannot handle long sequences effectively. 
	LSTM (Long Short-Term Memory) or bidirectional RNNs~\cite{Ma2017} can be trained using all available input information in the past and future, and have been used to alleviate the effect of the long sequence problem and improve the predictive performance. 
	Lipton et al.~\cite{Lipton2015} have been the first to successfully use LSTM to predict patients diagnosis, for multi-label classification.
	BEHRT~\cite{Li2019} is one of the most recent researches that provides the field with an accurate predictive model for the prediction of next diseases based on the transformer-based architecture in Natural Language Processing~\cite{Devlin2018}.
	
	Although fixed timesteps are perfectly suitable for many RNN applications, EHRs often contain event-driven and asynchronously sampled samples, in particular for the application considered in the present paper.
	Recent works try to capture time irregularity using recurrent neural networks. 
	Phased LSTM~\cite{Neil2016} tries to model the time information by adding a time gate to LSTM, which controls the update of the cell state, the hidden state and thus the final output.
	Another attempt is~\cite{Zhu2017time} where time gates are added to the LSTM in order to model time intervals and specifically better capture both of short-term and long-term sequential events.
	Time-aware LSTM~\cite{Baytas2017PatientSV} addresses time irregularity between two events in an LSTM architecture by decomposing the memory of the previous timesteps into short-term memory and long-term memory.
	Let us point out that, however, most of the works on EHR data cited above either ignores subsequence-level irregularity by partitioning the data into regular time windows and by aggregating the data within each window or handles this irregularity by simply adding the time-span as one coordinate of the features vector, see~\cite{Choi2015} for instance.

	\subsection{Aim of this paper and contributions}
	\label{sub: contributions}
	
	This paper aims at the construction of a predictive model for long-term and ``blurry'' relapses that occur after a medical act.
	We do not consider a short-term complication related to the act itself, but a long-term relapse which is observed only indirectly.
	In the example considered here, the medical act is a precise surgery, and the relapse is observed through longitudinal prescriptions of a specific set of drugs for urinary problems, over a period of 18 months after surgery.
	We use all the data available in SNIIRAM (French national health insurance information system, a huge database containing health reimbursements claims of almost all French citizens since 2015, see Section~\ref{sub: SNIIRAM} for details) for patients on which such a surgery has been performed (TURP surgery, see Section~\ref{sub: BPH}).
	
	\paragraph{The ZiMM model.}
	
	The first natural idea is to cast this problem as a classification problem for ``relapse'' versus ``no relapse'' after the medical act.
	However, such a naive approach cannot work here.
	Indeed, the definition of a binary label would require threshold choices, both about time and dosage: after how many time and amount of drugs prescribed do we consider that there is a relapse?
	Such thresholds might depend on various clinical practices, habits of the patient, and many other exogenous factors.
	In this work, we propose to work directly on the data observed, by using all the ``blurry'' observations at once to train an end-to-end architecture.
	For this purpose, we introduce a new methodological contribution, namely the ZiMM model (Zero-inflated Multinomial Mixture) which is described in Section~\ref{sub:zimm-model} below.

	\paragraph{The ZiMM Encoder-Decoder (ZiMM ED).}
	
	We introduce an end-to-end Encoder-Decoder architecture trained with an objective, based on the negative log-likelihood of the ZiMM model.
	All available patients' claims before the medical act are first encoded into a single embedding vector by the Encoder.
	This Encoder combines embedding layers, self-attention layers and recurrent layers to output an embedding vector of the full patient pathway prior to the medical act.
	This embedding is used as input to a Decoder which is dedicated to the learning of the parameters of the ZiMM model.
	This architecture is described in Sections~\ref{sub:zimm-encoder} to~\ref{sub:training} below.
	
	\paragraph{Specifics of this work.}
	
	The specificity of our approach lies in several points. 
	First of all, it is performed on non-clinical EHR data, and we consider long-term predictions (18 months ahead).
	Our end-to-end architecture addresses several challenges, such as variable-size inputs, confounding interactions between medical codes, long-term predictions.
	Finally, we do not inject any prior knowledge: the embeddings of patient pathways are fully trained in an end-to-end fashion, we do not use any strong preprocessing or simplifying aggregations, the data is used almost in its raw form.
	We keep the smallest granularity possible both on the codes (we use the actual medical codes instead of encompassing categories) and on time, namely we work on the original 1-day time scale.
	
	\paragraph{Organization of the paper.}
	
	The ZiMM model is described in Section~\ref{sub:zimm-model} while the ZiMM ED architecture is described in Sections~\ref{sub:zimm-encoder} to~\ref{sub:training}.
	Some details concerning data preprocessing are provided in Section~\ref{sub:preprocessing}.
	Section~\ref{sec:experiments} provides results and a comparison with strong baselines, together with some variations of the architecture to fine-tune the final performance and we conclude in Section~\ref{sec:conclusion}.

	\section{Proposed architecture}
	\label{sec:model}

	The ZiMM ED architecture has three main building blocks: the ZiMM (Zero-inflated Multinomial Mixture) model (see Section~\ref{sub:zimm-model}), an Encoder (see Section~\ref{sub:zimm-encoder}) and a Decoder (see Section~\ref{sub:zimm-decoder}), and is described in Figure~\ref{fig:architecture} below.
	Given a patient $i \in \{1, \ldots, n\}$ (among $n$ patients), with medical act of interest occurring at time $T^i$, the steps through this architecture are roughly as follows:
	\begin{itemize}
		\item All the medical codes (drugs, diagnosis, medical procedures) observed longitudinally in the life of patient~$i$ before $T^i$ are embedded, then aggregated in a time window using a self-attention mechanism. 
		The time distance (in days) to $T^i$ of each day with a non-empty set of codes is embedded as well, and we embed as well hospital stay durations.
		This sequence of vectors is then fed to a (or several) recurrent layers.
		This corresponds to Steps~1,2 and~3 in Figure~\ref{fig:architecture}. 
		This leads to an embedding vector $x_i \in \R^d$ of the full pathway of patient $i$ prior to $T^i$.
		\item The vector $x_i \in \R^d$ is concatenated with other static features of the patient, and used as the input of the Decoder, which is based on several recurrent layers, in order to produce the parameters of the ZiMM model. This corresponds to Step~4 in Figure~\ref{fig:architecture}.
	\end{itemize}
	Precise descriptions of all steps are provided below, starting with the ZiMM model, followed by some details concerning data preprocessing, and the Encoder-Decoder architecture.
	
	\begin{figure}[h!]
		\centering
		\includegraphics[width=0.9\textwidth]{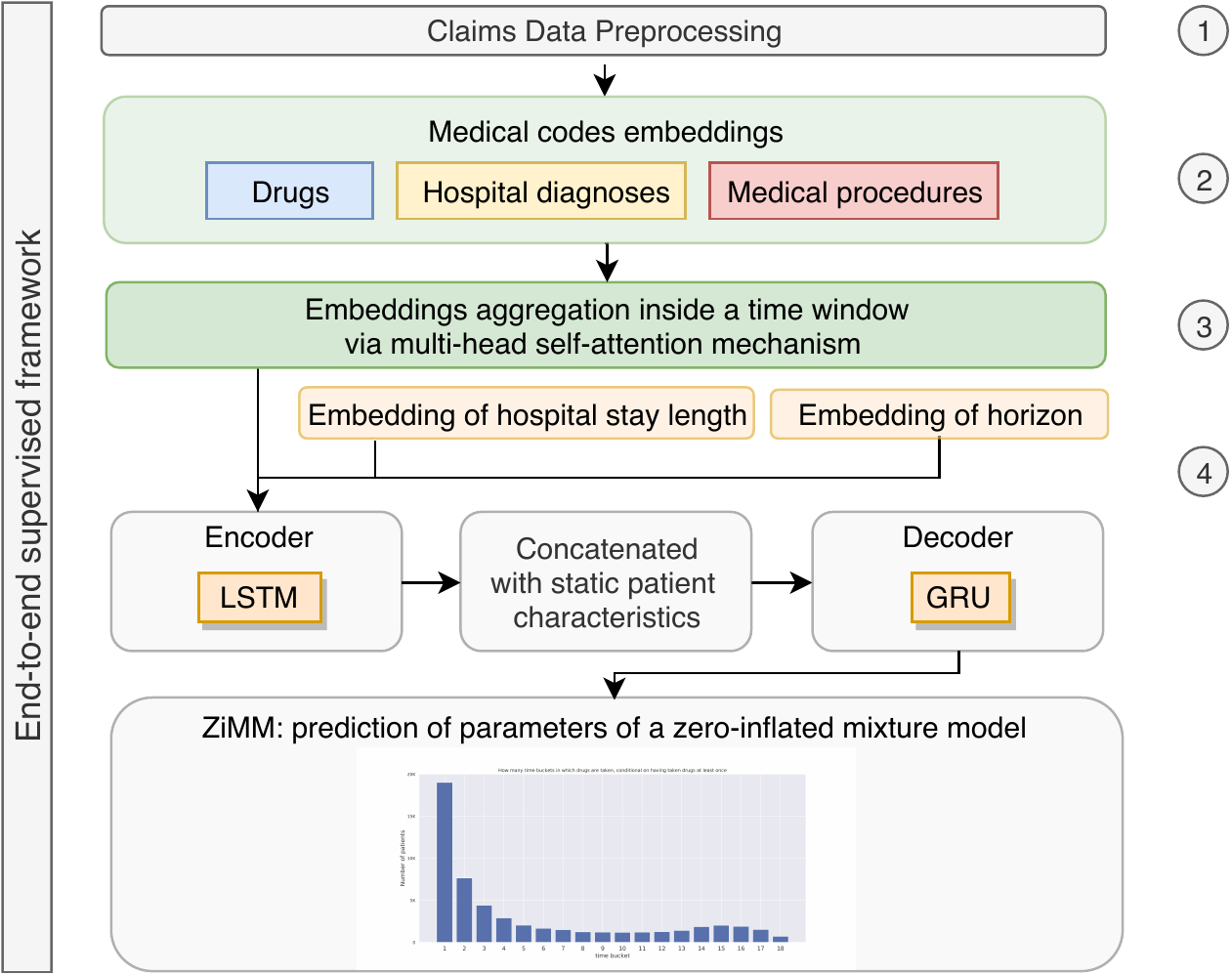}
		\caption{\small Architecture of the ZiMM Encoder-Decoder end-to-end architecture. A detailed graphical representation of Steps 2-4 is shown in Figure~\ref{fig:graphical_illustration} below.}
		\label{fig:architecture}
	\end{figure}

	\subsection{ZiMM: Zero-inflated Multinomial Mixture model}
	\label{sub:zimm-model}
	
	Assume for now that we have an embedding vector $x_i \in \R^d$ that encodes the full health pathway of patient $i \in \{1, \ldots, n\}$ prior to $T^i$.
	This vector is the output of the ZiMM-Encoder described in Section~\ref{sub:zimm-encoder} below.
	Moreover, we observe a vector of labels $y_i = [y_{i, 1}, \ldots, y_{i, B}] \in \N^B$, which corresponds to the blurry observation of the relapse after $T^i$. 
	Here, $B$ corresponds to the number of time intervals considered after $T^i$ and $y_{i, b} \geq 1$ is the number of blurry relapses observed in time bucket $b$, so that $y_{i, b} = 0$ means that no blurry relapse is observed in time bucket $b$.
	In the example considered Section~\ref{sec:experiments} below, $B=18$ corresponds to a 18-months period and $y_{i, b}$ is the number of drugs (among a set of drugs for urinary problems, see Section~\ref{sub:labels}), purchased by patient~$i$ in time bucket~$b$.
	
	We introduce $n_i = \sum_{b=1}^B y_{i, b} \in \N$, the overall number of blurry relapses of patient $i$.
	Let us point out that a binary classification problem ($n_i>0$ versus $n_i=0$) may wrongly classify very different situations corresponding to the same $n_i$.
	Consider for instance two extreme situations where first, we have $n_i = 1$ with $y_{i,1} = 1$ (and consequently $y_{i,b}=0$ for $b=2, \ldots, B$), and second, we have $n_i=1$ with $y_{i,B}=1$ (and consequently $n_{i,b} = 0$ for $b=1, \ldots, B-1$). 
	The first case can be due to exogenous factors, such as the patient simply kept buying the drug after the surgery just out of a pure habit (prescriptions can run for a long period), while in the second example, there is no doubt that a relapse is occurring.
	
	Therefore, we need to find a way to model the whole vector $y_i$, so that $n_i$ is not fixed and includes zero-inflation, namely a parametrized likelihood for $n_i = 0$.
	For that purpose, we introduce the following ZiMM model, where we suppose that $n_i \in \{ 0, 1, \ldots, B \}$ is distributed (conditionally to $x_i$) as
	\begin{equation}
		\label{eq:dist_n}
	\mathbb P(n_i = k | x_i) = \pi_k(x_i) \quad \text{for} \quad k \in \{0, \ldots, B\},
	\end{equation}
	where $\pi_b(x_i)$ are such that $\sum_{b=0}^B \pi_b(x_i) = 1$ and $\pi_b(x_i) \geqslant 0$ (coming out of a softmax activation for instance).
	These parameters correspond to the categorical distribution specific to patient $i$, and are constructed by the ZiMM-Decoder (see Section~\ref{sub:zimm-decoder} below) from $x_i$.
	Then, we assume that the distribution of $y_i$ conditional to $n_i=b$ and $x_i$ follows either a Dirac distribution on vector (of size $B$) $[0, \ldots, 0]$ whenever $b=0$, or a multinomial distribution of parameters $b$ and $p_{b, 1}(x_i), \ldots, p_{b, B}(x_i)$, namely
	\begin{equation}
		\label{eq:dist_y_cond_x_and_n}
	y_i |(x_i, n_i = b) \sim 
	\begin{cases}
	\delta_{[0, \ldots, 0]} \quad &\text{ if } b = 0, \\
	\text{Multinomial}(b, p_{b, 1}(x_i), \ldots, p_{b, B}(x_i)) \quad &\text{ otherwise},
	\end{cases}
	\end{equation}
	where $p_{b, 1}(x_i), \ldots, p_{b, B}(x_i)$ are the parameters of a multinomial distribution whenever $n_i=b$, for each $b=1, \ldots, B$.
	Once again, these parameters are specific to patient $i$ thanks to the embedding vector $x_i \in \R^d$, and are constructed by the ZiMM-Decoder.
	These parameters must satisfy $\sum_{b'=1}^B p_{b, b'}(x_i) = 1$ for $b=1, \ldots, B$, and $p_{b, b'}(x_i) \geqslant 0$ for any $b, b' = 1, \ldots, B$, which is easily achieved with a softmax activation applied row-wise.
	By combining~\eqref{eq:dist_n} and~\eqref{eq:dist_y_cond_x_and_n}, with end up with the following mixture distribution for $y_i$ conditionally on $x_i$:
	\begin{equation}
	y_i | x_i \sim \pi_0(x_i)  \delta_{[0, \ldots, 0]} + \sum_{b=1}^B \pi_b(x_i) \text{Multinomial}(b, p_{b, 1}(x_i), \ldots, p_{b, B}(x_i)).
	\end{equation}
	Zero-inflation corresponds to the case where $n_i = 0$ and has likelihood measured by $\pi_0(x_i)$.
	The ZiMM model has two sets of parameters that are functions of the feature vector $x$, namely
	$\{\pi_b(x)\}_{b \in \{ 1, \ldots, B\}}$ that parametrizes the distribution of the total number of blurry relapses and $\{p_{b, b'}(x)\}_{b, b' \in \{ 1, \ldots, B\}^2}$ that parametrizes the dynamics of these relapses.
	As explained in Section~\ref{sub:zimm-decoder} below, the dynamics of $\{p_{b, b'}(x)\}_{b' \in \{ 1, \ldots, B\}}$ for each $b$ are learned by dedicated recurrent layers in ZiMM-Decoder.

	\subsection{Preprocessing of the data}
	\label{sub:preprocessing}
	
\begin{figure}[h]
		\centering
		\includegraphics[width=\textwidth]{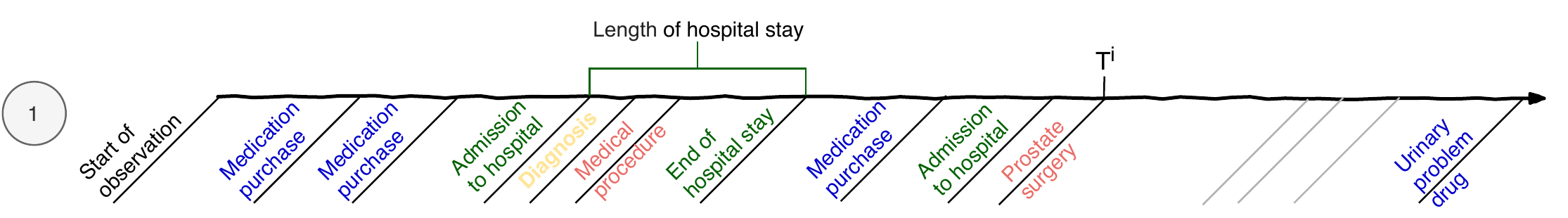}
		\caption{\small Illustration of the sequence of claims observed for a patient. This corresponds to the output of Step~1 from Figure~\ref{fig:architecture}.
			Through claims, we observe three types of events: drug purchases (blue), medical procedures (red) and diagnosis (yellow) before the medical act (prostate surgery in the example) that happens at time $T^i$ for patient $i$. All events are timestamped, and the time delta is a day long. Several events can occur the same day and some days have no event: events are typically very irregularly sampled.
			All these events, observed before $T^i$, are used to learn the embedding vector $x_i \in \R^d$ of patient $i$.
			After $T^i$, we only keep the events corresponding to the blurry relapses considered (drug purchases among a set of drugs for urinary problems). These blurry relapses are used to build the label vector $y_i \in \N^B$ of patient $i$.}
		\label{fig:patient_timeline}
	\end{figure}

	After a preliminary preprocessing based on our SCALPEL3 library described in~\cite{scalpel2019}, the data considered in this paper comes in the form of a table, where each row represents an event with the patient ID, the type of event (drug, diagnosis or medical procedure), its corresponding code, the start date and the end date of the event.
	Figure~\ref{fig:patient_timeline} provides an illustration of the sequence of claims observed for a single patient.
	If an event is related to a hospital stay, the start date is the first day of hospitalization while the end is the exit date from the hospital. 
	Otherwise, the start and end dates are the same. 
	The time delta is a day-long, and no aggregation is performed to keep the data as raw as possible.
	
	Multiple events can occur within the same day, and the precise ordering of such events does not carry any information.
	Thus, we shuffle at random within-day events, to avoid any bias that could occur from 
	the data collection mechanism (a similar strategy can be found in~\cite{ChoiY2016}).
	
	All the events observed through claims before $T^i$ are used as inputs to the Encoder described in Section~\ref{sub:zimm-encoder} below. 
	The output of the Encoder is the embedding vector $x_i \in \R^d$ of patient $i$.
	The blurry relapses observed after $T^i$ are used to build the label vector $y_i \in \N^B$ of patient $i$.

	\subsection{The ZiMM Encoder} 
	\label{sub:zimm-encoder}
	
\begin{figure}[h]
		\centering
		\includegraphics[width=0.9\textwidth]{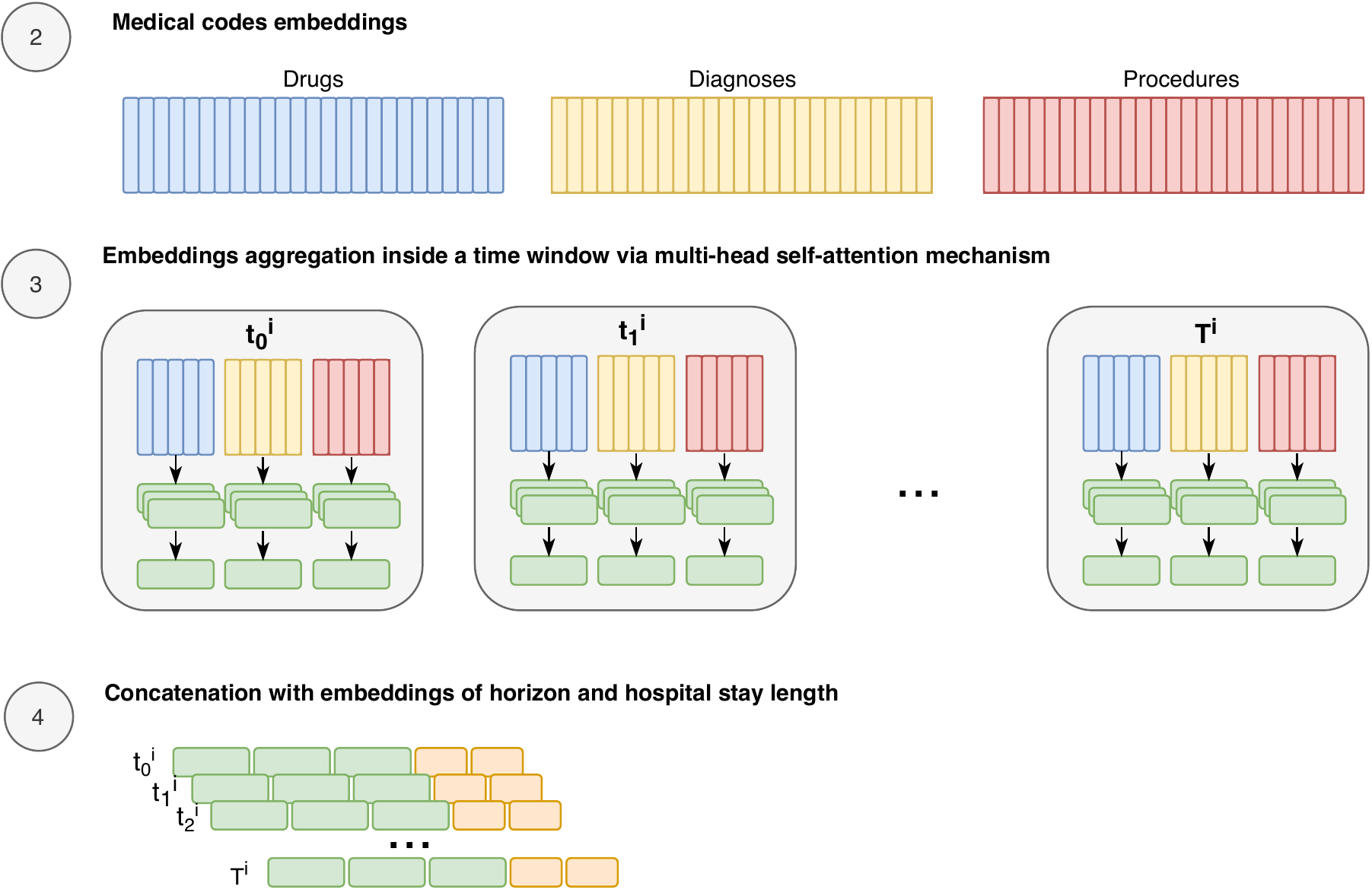}
		\caption{\small A graphical illustration of representation learning of patient pathway in the ZiMM Encoder: (2) embedding tensors for each type of medical code; (3) aggregation of embedding vectors through multi-head self-attention inside a time window, where each gray block corresponds to one day ($t_0^i, t_1^i, \ldots$), and $T^i$ is the day of the medical act of interest; (4) concatenation of aggregated embedding vectors with embeddings for time horizon and hospital length stay. }
		\label{fig:graphical_illustration}
	\end{figure}

	The ZiMM Encoder corresponds to Steps~(2) to~(4) in Figure~\ref{fig:architecture}, a more detailed illustration is shown in Figure~\ref{fig:graphical_illustration}.
	Let us provide now details about each step, following the flow of the data.
	
	\paragraph{Medical codes and timestamps embeddings.} 
	\label{sub:embeddings}
	
	The following is applied separately for each type of event code (drugs, medical procedures and diagnoses).
	Codes are first tokenized, and each unique token is individually mapped to an embedding vector in $\mathbb R^{d_E}$ which is learned during training, where $d_E$ is a hyper-parameter to be tuned (see Section~\ref{sub:results}).
	We consider only tokens that occur at least 50 times in the training dataset. 
	To address time irregularity between two asynchronously timestamped events, we do the following: for any event occurring at time $t \leq T^i$ in the observation period of patient $i$, we compute the "time horizon" as $T^i - t$, namely the distance (in days) between the event and medical act. 
	Moreover, whenever it makes sense, we compute $\text{end}-\text{start}$, the duration of the event, which corresponds to the duration of an hospital stay defined as the time between hospital admission and discharge.
	These two integers (in number of days) are also tokenized and replaced by a learned embedding vector.
	This allows the encoder to learn to put more or less emphasis on events that are close or far from $T^i$, and to exploit the duration of events as a proxy for severity of medical procedures and diagnoses.
	The patient age in years at $T^i$ is also similarly bucketized and embedded.
	The embedding of medical codes corresponds to Step~(2) of both Figures~\ref{fig:architecture} and~\ref{fig:graphical_illustration}.
	In our experiments, the default model uses $d_E = 64$ for the three types of codes and several regularization strategies are applied to avoid overfitting: weights decay ($\ell_2$ regularization with strength $10^{-2}$), multiplicative Gaussian noise, and layer normalization~\cite{Ba2016}.
	Since we observed empirically that these embeddings are prone to overfitting for the application considered in the paper, a comparison of the different regularization strategies is provided in Section~\ref{sub:results}.
	
	\paragraph{Embeddings aggregation through self-attention.} 
	
	Several events can occur at the same time (within the same day), so that the number of codes observed within a day is highly heterogeneous. 
	Moreover, such codes are not likely to contribute equally to the vector representation of the day, and their order is not informative.
	Therefore, we need to perform a trainable aggregation of these codes in order to produce a representation of the patient history at each timestamp.
	An approach can consist of using a hierarchy relationship between different diagnosis and treatment inside each patient visit~\cite{Choi2019}. 
	However, in the data considered here, diagnosis codes are not explicit each time there is treatment; in particular, drugs purchases are almost never related to a diagnosis code (unless they are related to a hospitalization).
	Hence, we use a self-attention mechanism~\cite{Vaswani2017, Lin2019} using a \textit{bag-of-features} approach to learn how to combine embedding vectors within the same day, following previous successful applications of self-attention for fusing disease embeddings~\cite{Luo2019}. 
	
	Let $C$ be a set of codes with cardinality $|E|$ (for drugs, medical procedures or diagnoses) and let $E_C \in \R^{d_E \times |C|}$ be the corresponding matrix of concatenated embedding vectors. 
	The multi-headed self-attention aggregation mechanism contains two layers, the first with $K$ heads, each of which is a self-attention function that generates a probability vector of size $|E|$ from \(E_{C}\):
	\begin{equation}
	\label{eq:self-attention-1}
	w_k = \softmax(\alpha_k^\top \tanh(A_k E_{C})),
	\end{equation}
	for $k = 1, \ldots, K$, where $\alpha_k \in \R^{1 \times d_E}$ and $A_k \in \R^{d_E \times d_E}$ are to be trained and the second layer outputs a $d_E$-dimensional vector given by
	\begin{equation}
	\label{eq:self-attention-2}
	E_C \mu_C \quad \text{where} \quad \mu_C = \softmax(b^\top \tanh(B W)),
	\end{equation}
	with $W = \text{concatenate}(w_1, \ldots, w_k) \in \R^{K \times |C|}$ and $b$ 
	and $B$ are trainable parameters. 
	We use the $\tanh$ squashing activation function following~\cite{Lin2019} and~\cite{Luo2019}, since in these layers it is considered good practice~\cite{lecun_efficient_2012} to produce centered activations that are bounded in $[-1, 1]$, to help the back-propagation process involved in the training of the model.
	This self-attention mechanism is trained and performed separately for each type of code, as displayed in Step~(3) of Figure~\ref{fig:graphical_illustration}.
	Note that this aggregation approach is permutation-invariant for codes that occur within the same day. 
	We use dropout, drop-connect~\cite{Wan2013}, Frobenius-norm weight penalization from~\cite{Lin2019} and weight-decay to regularize this self-attention mechanism.

	\paragraph{Health pathway encoder.}
	
The inputs of the health pathway encoder, at each timestamp, contains the following embedding vectors, that are concatenated together:
	\begin{itemize}
		\item Medical code embedding computed by Equations~\eqref{eq:self-attention-1} and~\eqref{eq:self-attention-2};
		\item Time-horizon embedding (see above);
		\item Duration of event embedding (see above).
	\end{itemize}
	For each patient, this leads to a sequence of fixed-sized embedding vectors that encode both medical and time information at each (non-empty) timestamp $t \leq T^i$, as displayed in Step~(4) of Figures~\ref{fig:architecture} and~\ref{fig:graphical_illustration}.
	Now, this sequence of vectors is used as the input of a stack of layers, including recurrent layers (LSTM~\cite{hochreiter1997long}, bi-directional LSTM (bi-LSTM)~\cite{schuster1997bidirectional}, GRU~\cite{cho2014learning}) or convolutional layers, in a \emph{sequence-to-one} network architecture, since we want to output a single vector $x_i \in \R^d$ that encodes the full pathway of a patient before $T^i$.
	The results obtained with different types of layers are shown in~Section~\ref{sub:results} below.
	We use again dropout on the inputs and on the recurrent units to prevent overfitting.
	Finally, the output vector of the encoder is concatenated with an embedding vector of the age of the patient, leading to the final vector $x_i \in \R^d$ that encodes the full pathway of patient $i$ before $T^i$.
	This vector is the input of the ZiMM Decoder described in the next Section.

	\subsection{The ZiMM Decoder} 
	\label{sub:zimm-decoder}
	
	The decoder uses the input vector $x_i \in \R^d$ of patient $i$ to construct the parameters of the ZiMM model, and the whole architecture is trained against the negative log-likelihood of the ZiMM model from Section~\ref{sub:zimm-model} computed at the label vector $y_i = [y_{i, 1}, \ldots, y_{i, B}]$ containing the blurry relapses.
	The parameters $\{ \pi_b(x_i) \}_{b=1 \ldots, B}$ and $\{ p_{b, b'}(x_i) \}_{b, b' \in \{ 1, \ldots, B\}^2 }$ are highly dependent and are time-ordered, so that a specific architecture is used to model these dependencies.
	The mixture probabilities are learned through a fully connected feed-forward layer (FFN) that use as input $x_i$, namely
	\begin{equation}
	\pi_0(x_i), \ldots, \pi_B(x_i)= \text{softmax}(\text{FFN}(x_i)),
	\end{equation}
	and another recurrent layer (RNN) uses as well $x_i$ in order to produce hidden states given by
	\begin{equation}
	\label{eq:decoder_hidden}
	h_t = \mathrm{RNN}(h_{t-1}, x_i)
	\end{equation}
	for $t=1, \ldots, B$.
	Then, we use different recurrent layers ($\text{RNN}_b$) in parallel for each value $n_i = b \in \{1, \ldots, B\}$ using 
	\begin{equation}
	h^b_t = \mathrm{RNN}_b(h^b_{t-1}, h_t)
	\end{equation}
	for $t=1, \ldots, B$, since the multinomial distributions vary strongly conditionally on $n_i = b$.
	Finally, a softmax activation is applied on $h^b_t$ along $t=1,\ldots, B$ to produce the parameters $p_{b, 1}(x_i), \ldots, p_{b, B}(x_i)$.
	Details on the type of layers used and the tuning of hyperparameters is provided in Section~\ref{sec:experiments} below.
	
	\subsection{Training} 
	\label{sub:training}
	
	The full ZiMM Encoder-Decoder architecture is trained in an end-to-end fashion by minimizing the average negative-log likelihood over all patients $i=1, \ldots, n$:
	\begin{equation}
	\ell_i(\Theta) = \log(\pi_0(x_i)) \1_{n_i=0} + \sum_{b=1}^B \Big[ \log(\pi_b(x_i)) + \log \Big( \frac{b!}{\Pi_{b'=1}^B y_{i,b'}!} \Big) + \sum_{b'=1}^B y_{i, b'} \log(p_{b, b'}(x_i)) \Big] \1_{n_i=b},
	\end{equation}
	where $\Theta$ stands for the concatenation of all the trainable parameters involved in the layers described in Sections~\ref{sub:zimm-encoder} and~\ref{sub:zimm-decoder}.
	The choices of optimizer, learning rate schedule and other hyperparameters are described in Section~\ref{sec:experiments}.

	\section{Application: prediction of post-surgical relapse of urinary problems}
	\label{sec:experiments}
	
	In this section, we apply the ZiMM ED architecture to predict the blurry relapse of urinary problems after a TURP surgery for patients with benign prostatic hyperplasia (see Section~\ref{sub: BPH} below) using a cohort constructed from the French SNIIRAM database (see Section~\ref{sub: SNIIRAM} below).
	Then, we explain in details the way the labels are built (Section~\ref{sub:labels}), the different steps involved in the cohort construction (Section~\ref{sub:cohort}), the evaluation metric and implementation details (Section~\ref{sub:evaluation}).
	Then, the remaining of this section describes the baselines and a comparison with ZiMM ED, followed by an ablation study.

	\subsection{SNIIRAM: A non-clinical claims dataset}
	\label{sub: SNIIRAM}
	
	SNIIRAM (French national health insurance information system) contains health reimbursements claims of almost all French citizens since 2015 (more than 65 million) and has been used for health research on many topics, to cite but a few~\cite{Scailteux2019, Atramont2018, Fonteneau2017}.
	Since, in France, most health-cares are at least partially reimbursed by the administration, 
	this database contains records corresponding to very various information going from hospital stays to drug purchases in city pharmacies, all coded with different systems~\cite{Tuppin2017}. 
	Absence of clinical information and ``forced'' normalization makes this data highly complex and heterogeneous. 
	A consequence is that it requires a lot of domain expertise about the way healthcare is reimbursed in France in order to prepare it as training data for machine learning algorithms.
	In this paper, we exploit medication, procedure and diagnosis codes only.
	Diagnoses are primarily coded with ICD-10, while procedures are coded with CCAM, a French medical classification of clinical procedures. Medications codes use the French pharmaceutical categories CIP13~\cite{Bezin2017} that we map to the international ATC (Anatomical Therapeutic Chemical Classification) classification system. 
	
	Let us stress that while this data is extremely rich and almost population-wide over France, it is not clinical data, but only claims data, clinical information being only latent in the codes appearing in reimbursement information. 
	No vital bedside information, lab tests results or natural language notes from clinicians is available.
	An important amount of data preprocessing and domain expertise is required for data extraction, in particular in order to clean existing inaccuracies in codes and timestamps.
	This is achieved through our SCALPEL3 library~\cite{scalpel2019}, which is a separate topic of research that this paper builds upon.
	Furthermore, the identification of a disease state (indication, concomitant disease or outcome) does not rely only on a single source of information but on the convergence of information from different sources. This might include, for instance, the presence of a chronic disease registration, hospital diagnoses, tracer drugs, and procedures or lab tests~\cite{Bezin2017}, that are all observed in the data through claims.
	
	We believe that these downsides are counter-balanced by the exhaustive nature of this data. It is exhaustive both in terms of population (it includes nearly all the population living in France) and in terms of healthcare events. Indeed, in France, almost all healthcare spendings are at least partially reimbursed by the government, so almost each healthcare of each individual is associated to at least one event in the database. 
	This makes it a very rich and powerful database, with a relatively small bias, 
	and many works have already proved successful in improving population-level health~\cite{Morel2019, Scailteux2019, Atramont2018, Fonteneau2017}.
	Though most EHR studies are dealing with clinical EHR's~\cite{article}, it is clearly crucial to exploit such claims-only EHR, and we hope that our work is a first step towards broader use, in healthcare and machine learning research community, of non-clinical claims datasets.
	
	Let us note that, since 2016,  the access to this database, for public interest research, is possible through the SNDS access pipeline~\cite{SNDS2019}. 
	It mainly offers access through the SAS software. An access using classical open source big data and AI frameworks (e.g. R, Python, Spark) including the SCALPEL3 library will be possible, by the end of 2020, through the Health Data Hub~\cite{hdh}, a 80m\$-funded national project which aims to be the national unique gateway to most French health data for operating public interest research (operated by both public or private entities) on modern infrastructures.

	\subsection{Benign prostatic hyperplasia (BPH)}
	\label{sub: BPH}
	
	BPH is a common urological disease that affects aging men all over the world \cite{Silva2014, Dahm2017}. 
	It causes urinary tract obstruction due to the unregulated growth of the prostate gland, causing lower urinary tract symptoms (LUTS)~\cite{Kim2016}. 
	The options for management of BPH include watchful waiting, pharmacotherapy, transurethral resection of the prostate (TURP), and other minimally invasive surgical treatments (MISTs) and open prostatectomy.
	Most studies consider TURP as the gold standard for surgical management of BPH~\cite{Hashim2015}. 
	Patients suffering from prostatic hyperplasia regularly take drugs for urination problems. Successful surgery should cure such problems so that patients should not need to continue their drugs for urination problems. However, it is often observed that patients retake such drugs after surgery. In some cases, it merely comes from a habit of taking routine drugs, but in many cases, it is related to a persisting urination problem~\cite{Macey2016, Chung2018}.
	
	An important problem, from a clinical point of view, is, therefore, to predict the outcome of such a surgery, on a long-time period following it (18 months is considered here), to improve the decision making of the clinicians, in particular, to help decide when surgery should be performed.
	Use of long-term data after TURP surgery is very scarce in the literature, with only a few studies available~\cite{Cornu2015}. 
	A recent analysis of 20 contemporary randomized clinical trials with a maximum follow-up of five years reports that TURP resulted in a significant improvement of maximum flow-rate and quality of life~\cite{Cornu2018, Lourenco2010}. 
	A second prostatic operation (re-TURP) has been reported at a constant annual rate of approximately 1-2\%. A review analyzing 29 RCTs found a re-treatment rate of 2.6\% after a mean follow-up of sixteen months~\cite{Cornu2018}. 
	In a large-scale study of 20,671 men, the overall re-treatment rates (i.e., either re-TURP, urethrotomy or bladder neck incision) were 5.8\%, 12.3\%, and 14.7\%, respectively at one, five, and eight years follow-up. 
	More specifically, the respective incidence of re-TURP was 2.9\%, 5.8\% and 7.4\%~\cite{Madersbacher2005}. 
	However, urology guidelines highlight the lack of extended follow-up after the surgery and no clear evidence explaining re-treatment.
	
	Building a model with the ability to predict the outcome of this surgery is of primary importance for improving population-level health, especially since prostatic problems are a common condition for aging men. 
	In this section, we use ZiMM ED to train such a model using a cohort based on SNIIRAM. 
	The outcome of this surgery is evaluated by the distribution of the blurry relapses, observed through the use of specific drugs related to urination problems.
	
	\subsection{Surgery identification and labels construction}
	\label{sub:labels}
	
	The identification of a TURP surgery in SNIIRAM is made through some specific CCAM codes provided by clinicians\footnote{We considered that a TURP surgery corresponds to the CCAM codes JGFA005, JGFA009, JGFA015, JDPE002 or JGNE003.}. 
	However, it happens that two TURP surgeries can occur in a pretty small amount of time. 
	This is likely to correspond to a case where a surgery has clearly failed and that a second surgery is required quickly, which is not the type of complication we are interested in.
	If the amount of time between the two surgeries is small, it would be improper to say that it corresponds to a relapse. 
	We thus define, following clinicians recommendations, a six week period after the first surgery, as a \emph{single surgery block}. 
	If another TURP surgery occurs inside the same block, we consider that it is part of the same medical act. 
	The timestamp of the event corresponds to the last surgery within the block, and provides the value of $T^i$ for any patient $i$.
	
	As far as the labels are concerned, a simple but efficient way to identify whether the urinary problems have not ceased or reappeared after a while is to see if the patient, at some point after the surgery, needs to take medications for these urination problems again\footnote{Following clinicians recommendations, we do not consider TURP surgeries that occur after the first surgery block, and consider only drugs prescriptions as blurry relapses.
		Consequently, we remove patients with repeated surgeries that are not part of the same block from the cohort construction, as explained in Section~\ref{sub:dataset}.}.
	For that purpose, we use a list (provided by clinicians) of 136 CIP-13 drugs that are mainly related to urination problems. 
	We choose to drive the prediction on a 18-months period after $T^i$. 
	So, using the notation of Section~\ref{sub:zimm-model}, we chose the number of buckets $B=18$ and bucket size of 30 days (540 days total).
	
	\subsection{Cohort construction and exploration}
	\label{sub:dataset}
	\label{sub:cohort}
	
\begin{figure}[h]
		\centering
		\sffamily
		\footnotesize
		\begin{tikzpicture} [auto,scale=0.75, every node/.style={scale=0.85},
		block_large/.style ={rectangle, draw=black, thick, fill=red!10,
			text width=22em, text centered, minimum height=4em},
		block_center/.style ={rectangle, draw=black, thick, fill=white,
			text width=8em, text centered, minimum height=4em},
		block_left/.style ={rectangle, draw=black, thick, fill=yellow!15,
			text width=18em, text ragged, minimum height=4em, inner sep=6pt, rounded corners=3mm},
		line/.style ={draw, thick, -latex', shorten >=0pt}]
		\matrix [column sep=5mm,row sep=3mm] {
			\node [block_large] (initial) {Patients covered by French universal health insurance with at least one urination problem event (surgery or medication) between 2010/01/01 and 2014/12/31 \\ (\textbf{$n$ = 5 136 308})}; 
			& \node [block_left] (excluded1) {Excluded ($n$ = 1 045 234): 
				\begin{enumerate}
				\item Age < 18; and > 110 if death date is missing ($n$ = 39 300)
				\item Women ($n$ = 1 005 933)
				\item Wrong hospital stays dates ($n$ = 1)
				\end{enumerate}
			}; \\
			\node [block_large] (men) {Exported cohort \\ (\textbf{$n$ = 4 091 074})}; \\
			\node [block_large] (cohortPHS) {Patients with surgery event between 2010/01/01 and 2013/06/30 \\ (\textbf{$n$ = 231 747})}; 
			& \node [block_left] (excluded2) {Excluded (n = 92 771):
				\begin{enumerate}
				\item More than 1 surgery bloc  ($n$ = 9 973)
				\item Death before $T$ + 18 months ($n$ = 14 617) 
				\item Less than 500 observation days in the 18-months follow-up period ($n$ = 68 181)
				\end{enumerate}
			}; \\
			\node [block_large] (analysed) {Analysed cohort \\ (\textbf{$n$ = 138 976})}; \\
		};
		\begin{scope}[every path/.style=line]
		\path (initial)   -- (excluded1);
		\path (initial)   -- (men);
		\path (men)       -- (cohortPHS);
		\path (cohortPHS) -- (excluded2);
		\path (cohortPHS) -- (analysed);
		\end{scope}
		\end{tikzpicture}
		\captionof{figure}{\small Flowchart leading to the final cohort considered in the experiments. The number $n$ stands for the number of patients remaining at each stage of preprocessing.}
		\label{fig:flowchart}
	\end{figure}
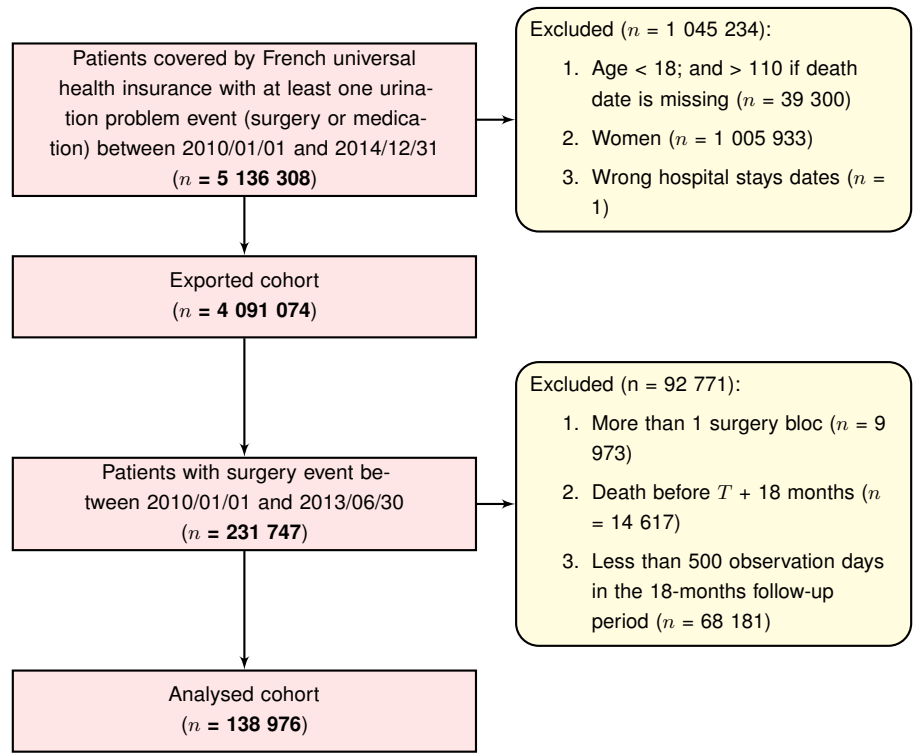

	The construction of the cohort follows the flowchart illustrated in Figure~\ref{fig:flowchart}.
	We start by selecting all SNIIRAM patients alive, with at least one medical event related to urination problem (observed through medication or surgery codes) between 2010/01/01 and 2015/12/31. 
	This selection gathers a little bit more than 5 million patients.
	Then, we keep only male patients over 18 years old on 2014/12/31. 
	This reduces the size of the cohort by roughly 1 million patients.
	Keeping only patients with a TURP surgery between 2010/01/01 and 2013/06/30 and with at least 2 events (whatever the types) occurring in the follow-up 18-months period leads to a cohort containing 231 747 patients.
	
	As explained in Section~\ref{sub:labels}, we compute surgery blocks (any surgery that occurs within a six week period after the first surgery is in the same block), that are considered as a single event of surgery. 
	We remove all the patients that have another surgery out of this first block.
	We choose a follow-up period of 540 days, corresponding to 18 buckets of 30 days, so that the \emph{follow-up} period is of the order of 18 months.
	Indeed, clinicians expect that such a relapse should occur not long after the first year following surgery.
	As shown in Figure~\ref{fig:flowchart}, the final cohort has 138 976 patients. 
	
	The numbers of medical codes, namely medications, procedures and diagnoses observed in the final cohort, are displayed in Table~\ref{tab:code-counts}.
	We count the number of unique codes, the number of codes observed at least 50 times and the number of codes remaining when keeping only the ones prior to $T^i$ for each patient~$i$.
	The statistics reported in Table~\ref{tab:code-counts} below and in all figures from this section use the raw coding scheme (that use CIP-13 for drugs), in order to better describe the statistics of the data, while for training ZiMM ED we replace CIP-13 by ATC (for drugs) since, as illustrated in Section~\ref{sub:results} below, it allows to reduce the vocabulary size of drugs while keeping the same predictive performance.
	As for subsequent encoding, we consider only tokens that occur at least 50 times in the training dataset, so that the encoder will see a total of 9 271 medical codes.
	\begin{table}[h]
		\centering
		\small
		\begin{tabular}{rcccc}
			\hline
			& Medications & Medical procedures & Diagnoses & Total \\ 
			\hline
			\#Unique  & 12 785 & 9 578 & 5 885 & 28 248 \\
			\#Unique before $T^i$ & 10 664 & 7 460 & 4 563 & 22 687 \\
			\#Unique seen $\geq 50$ times before $T^i$ & \textbf{6 713} & \textbf{1 155} & \textbf{1 403} & \textbf{9 271} \\ 
			\hline
		\end{tabular}
		\caption{\small Number of medical codes observed and remaining when applying the filters used in the cohort construction. Bold values correspond to the vocabulary sizes of the medical codes used when training the ZiMM ED architecture.}
		\label{tab:code-counts}
	\end{table}
	
\begin{figure}[h]
		\centering
		\includegraphics[width=0.5\textwidth]{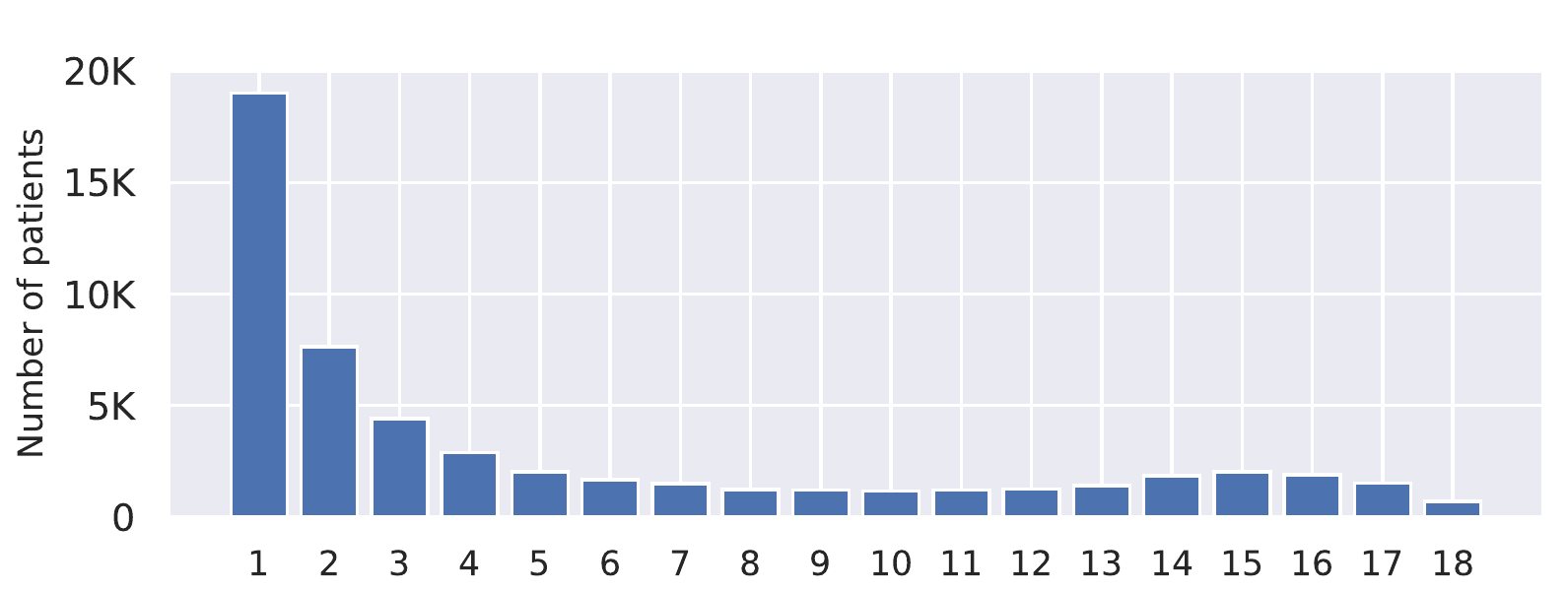}%
		\includegraphics[width=0.5\textwidth]{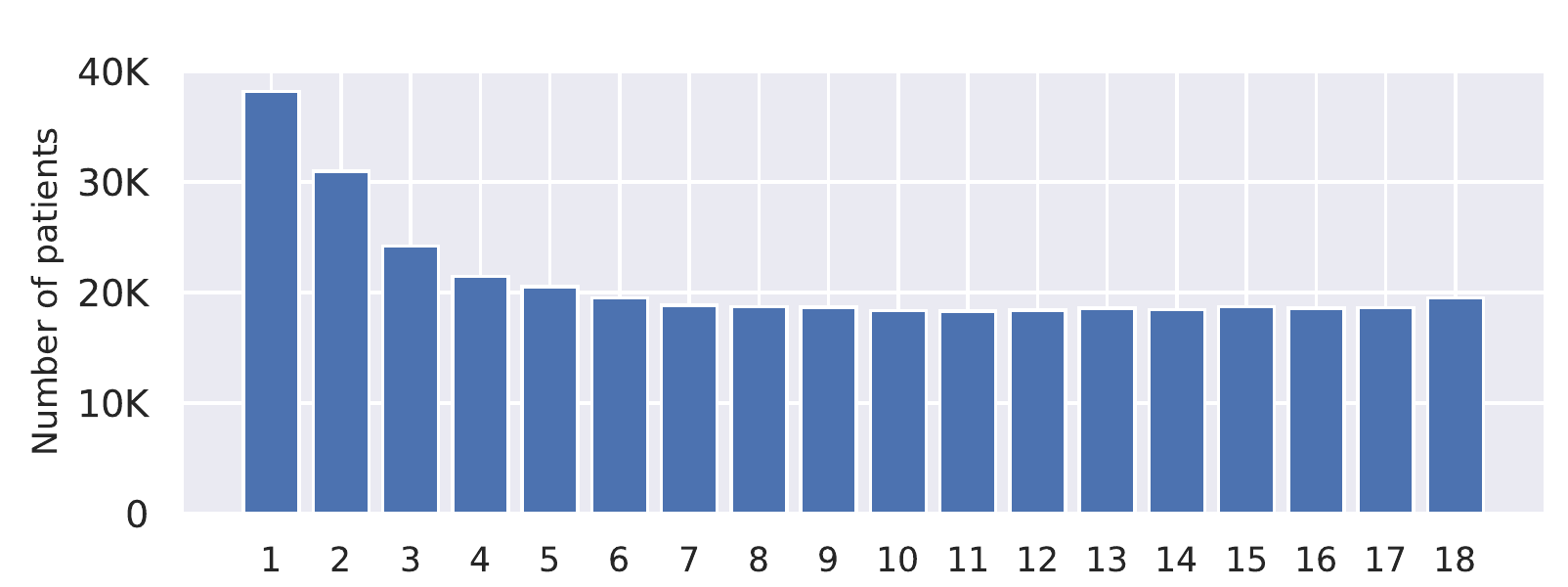}
		\caption{\small \emph{Left-hand side.} Number of patients as a function of the overall number of drugs prescriptions related to urinary problems during the follow-up 18 months period. The $y$-axis displays $\#\{ i : n_i = b \}$ where $b=1, \ldots, B$ is given by the $x$-axis.
			We observe a sharp decrease from the mode $b=1$ followed by a second weak mode around $b=15$. 
			The number of patients with $n_i = 0$ is equal to 84 328 and is not displayed for readability of the plot.
			\emph{Right-hand side.} Number of patients having at least one drug prescription (related to urinary problems) in each time bucket: the $y$-axis displays $\#\{i,~y_{i,b}\ge 1\}$ as a function of $b$ given by the $x$-axis.
			We observe a sharp decrease from the mode $b=1$ and a plateau between $b=7$ and $b=18$, since patients often continue to purchase the drug after the surgery.}
		\label{fig:distribution_ni}
	\end{figure}

	On the left-hand side of Figure~\ref{fig:distribution_ni}, we show, for this final cohort, the distribution of $n_i$, namely the number of patients with a given total number of drugs prescriptions (related to urinary problems) during the follow-up 18 months period.
	Namely, the $y$-axis of Figure~\ref{fig:distribution_ni} (left-hand side) displays $\#\{ i : n_i = b \}$ where $b=1, \ldots, B$ is given by the $x$-axis.
	We observe a sharp decrease for $b = 1, \ldots,5$ from the mode $b=1$, which corresponds to patients who continue to buy their drugs and just keep doing it (for several months) after their surgery.
	However, we observe a second mode around $b=15$ which certainly corresponds to the relapse we are interested in.
	This plot shows that the considered problem is in a ``weak signal'' setting, since the second mode is very small compared to the first (at $b=1$), and that the prediction of this relapse requires a dedicated methodology indeed.
	This observation is corroborated by the right-hand side of Figure~\ref{fig:distribution_ni}, which displays the number of patients having at least one drug prescription (related to urinary problems) on a specified bucket, namely the $y$-axis displays $\#\{i,~y_{i,b}\ge 1\}$ as a function of $b$  given by the $x$-axis.
	Once again, the function is decreasing from the mode $b=1$ and plateaus between $b=7$ and $b=18$, because of patients who continue to purchase the drug after the surgery.
	We do not observe a second mode as we did in the left-hand side, since the relapse we are interested in is blurry: the drug purchases that define this blurry relapse are heterogeneously distributed among patients and end up flattened out when aggregated in the display of the right-hand side.
	This motivates, once again, the dedicated methodology proposed by the ZiMM model.
	
\begin{figure}[h]
		\centering
		\includegraphics[width=0.47\textwidth]{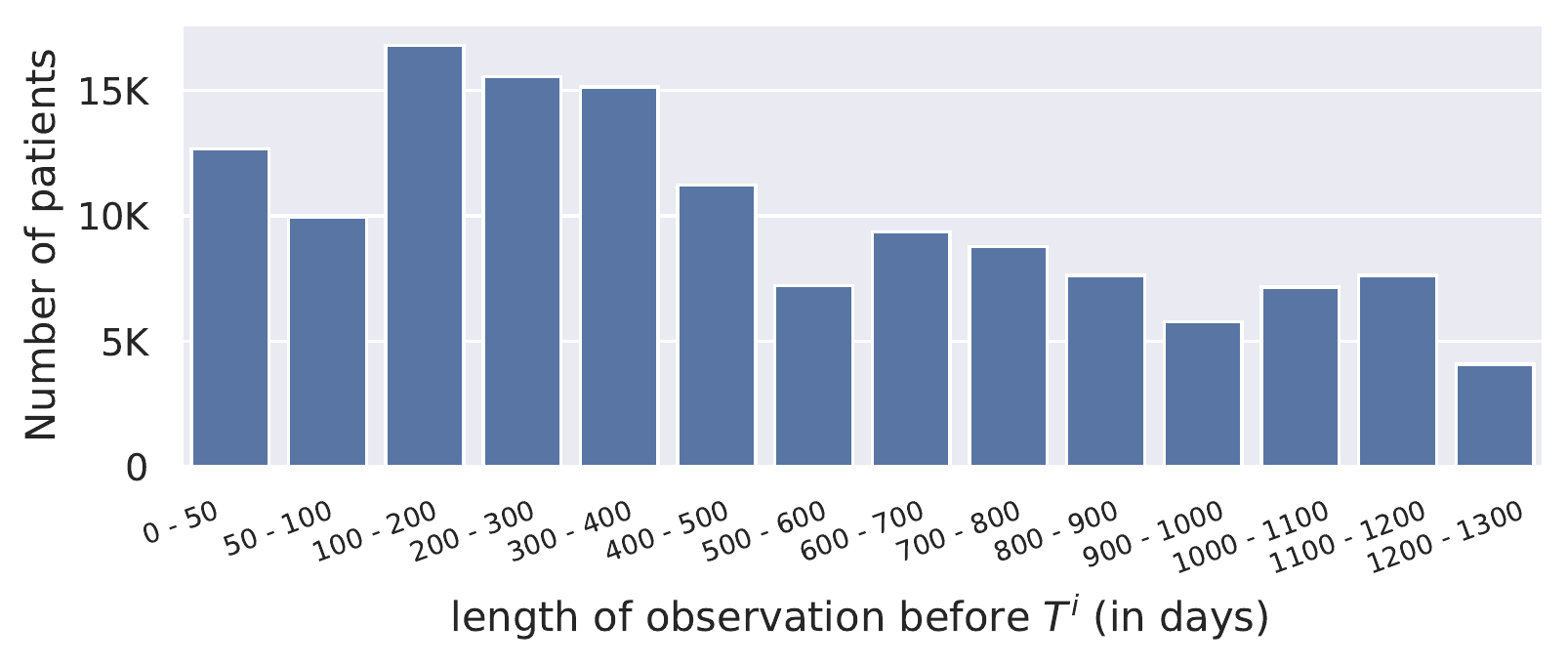}%
		\includegraphics[width=0.53\textwidth]{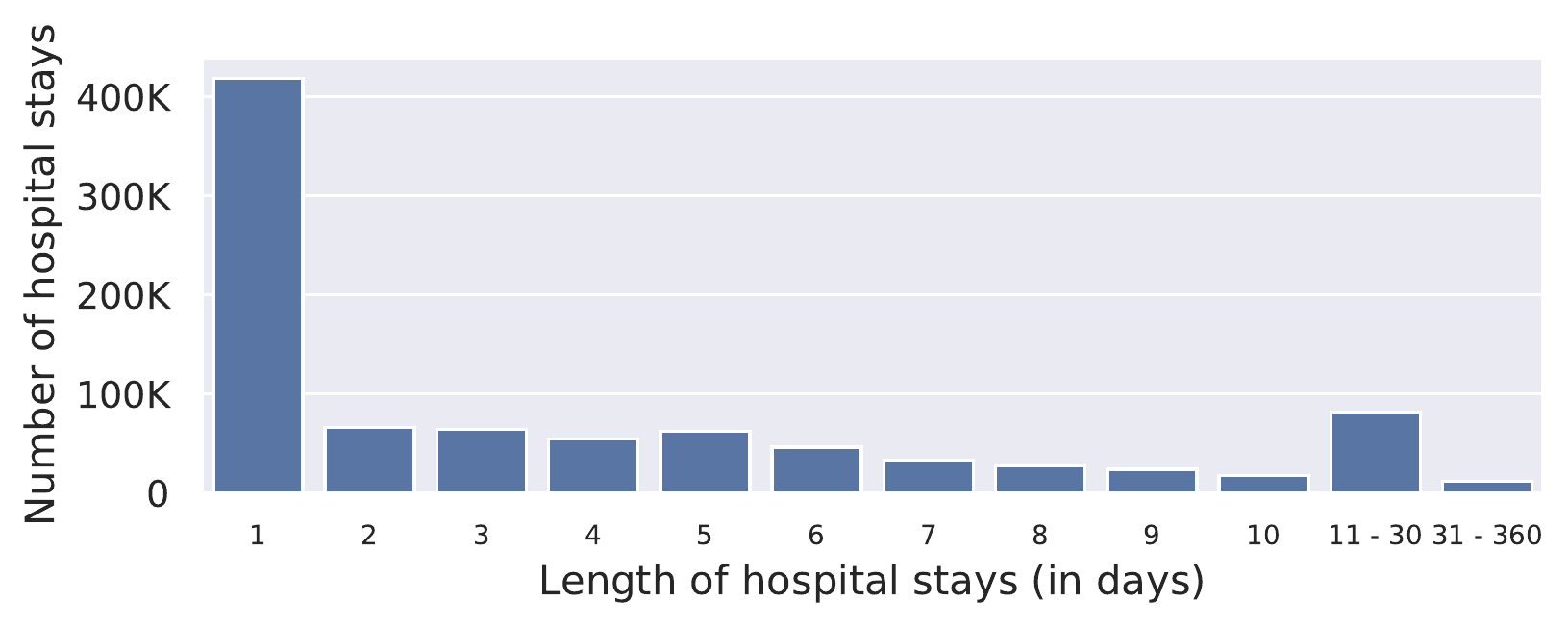}
		\caption{\small \emph{Left-hand side.} Distribution of the observation period of each patient before $T^i$. The strong variability of the observation period displayed here is related to the variability of the medical practice itself: some patients are treated for urological problems with drugs for a long time before the surgery, while other patients have surgery sooner.
			\emph{Right-hand side.} Distribution of the durations of all hospitalizations before $T^i$. Most hospitalizations are one-day short, and the distribution is heavy-tailed, since these hospitalizations can be related to a large set of possible health problems.
		}
		\label{fig:distribution_observation_period}
	\end{figure}

	In the left-hand side of Figure~\ref{fig:distribution_observation_period}, we show the distribution of the observation period of each patient before $T^i$. 
	We observe that it is highly heterogeneous: the maximum is reached at 1 274 days while the average is 486 days, and some patients have a very short observation period.
	This variability is related to the variability of the medical practice itself: some patients are treated for urological problems with drugs for a long time before the surgery, while other patients have surgery sooner.
	On the right-hand side of Figure~\ref{fig:distribution_observation_period}, we show the distribution of the durations of all hospitalizations before $T^i$.
	We observe that most hospitalizations are one-day short, and that the distribution is heavy-tailed, since these hospitalizations can be related to a large set of possible health problems, leading to very heterogeneous durations.
	Let us stress that we are considering \emph{all} hospitalizations that occur before $T^i$, and not only the ones related to a surgery or a urological problem, since all the health events of a patient $i$ are kept before $T^i$.
	
\begin{figure}[h!]
		\centering
		\includegraphics[width=0.51\textwidth]{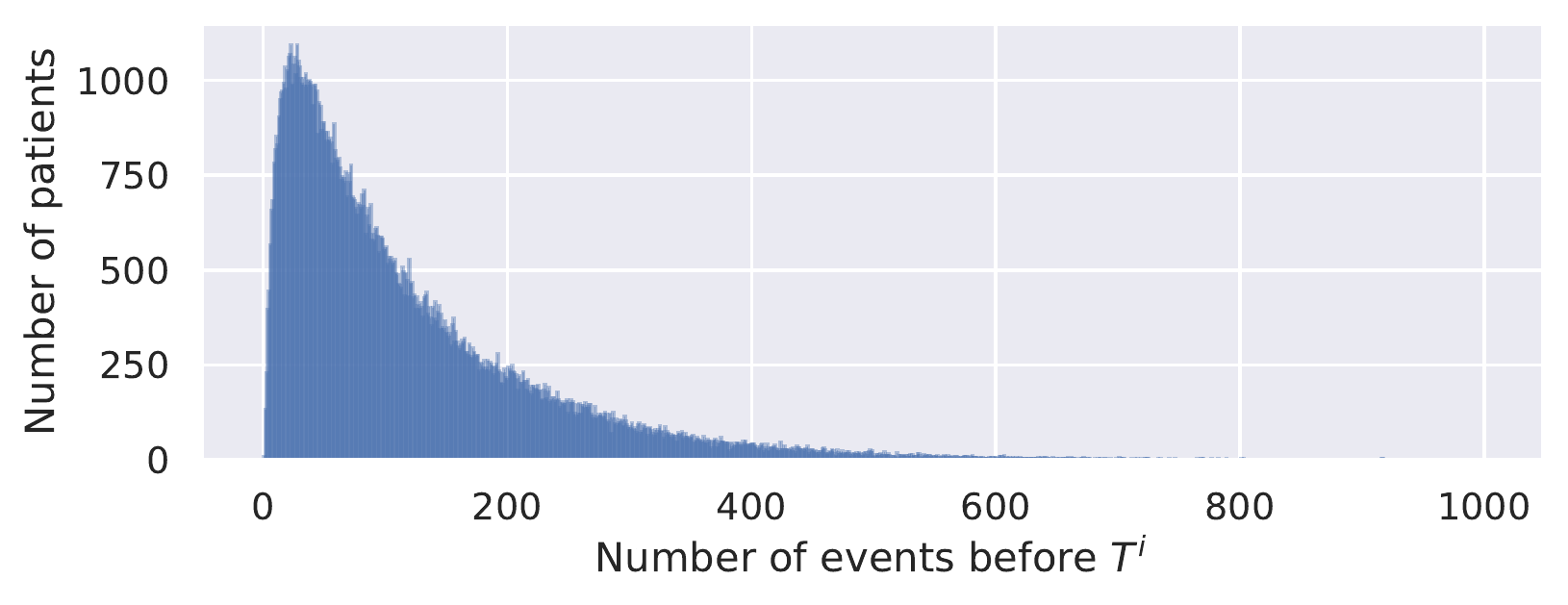}
		\includegraphics[width=0.48\textwidth]{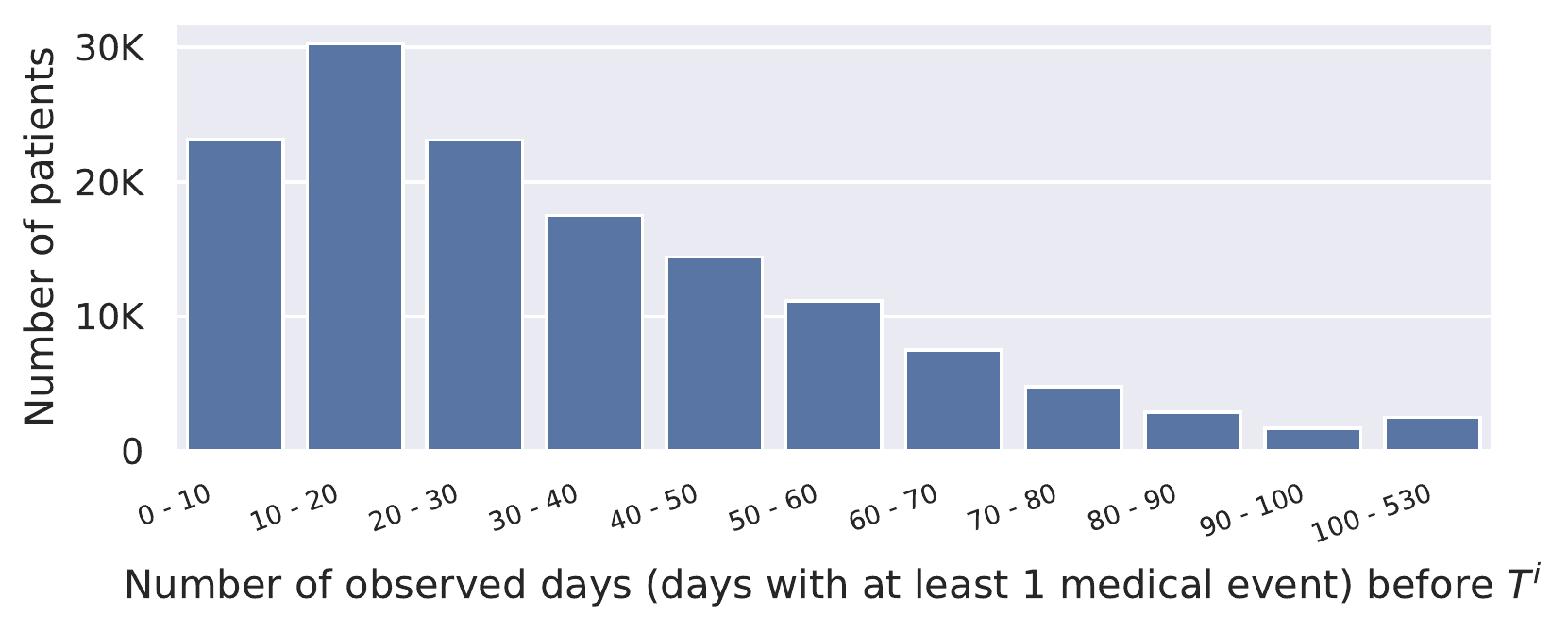}
		\caption{\small \emph{Left-hand side.} Distribution of the number of events per patient before $T^i$. The bulk of the distribution is between 5 and 200 unique events, with an average of 117 events per patient. \emph{Right-hand side.} Distribution of the number of days with at least one medical code in the history of patients before $T^i$. We observe that most patients have more than 10 days in their history with medical codes. Both figures show that most patients have a significant amount of medical code variability and time variability before $T^i$.}
		\label{fig:nb_events_before_surg}
	\end{figure}

	In the left-hand side of Figure~\ref{fig:nb_events_before_surg}, we display the distribution of the number of unique events per patient.
	We observe that the bulk of the distribution is between 5 and 200 unique events, with an average of 10 unique events.
	The right-hand side of Figure~\ref{fig:nb_events_before_surg} shows the number of days with at least one medical code in the history of patients before $T^i$. 
	We observe that most patients have more than 10 days in their history with medical codes. 
	Both displays from Figure~\ref{fig:nb_events_before_surg} show that 
	the health pathways of most patients before $T^i$ contain enough variability when assessed by the number of distinct codes, and the number of distinct days with at least one medical code, corresponding to a sufficient amount of variability to carry information for the prediction of the blurry relapse after $T^i$.

	\subsection{Implementation and evaluation metrics}
	\label{sub:evaluation}
	
	Our models are implemented with TensorFlow~2~\cite{tensorflow2015-whitepaper}.
	The ZiMM ED architecture is open-source in GitHub\footnote{\url{https://github.com/stephanegaiffas/zimm.git}}.
	All the models are trained on a machine equipped with 4 GeForce GTX 1080Ti GPUs and another machine with 3 Tesla V100 GPUs.
	All the models are trained with the Nadam optimizer~\cite{Kingma2014, Dozat2016} with learning rate 0.001.
	The hyper-parameters of all the models are selected through an extensive random grid search, whereas in order to reduce computation time, some hyper-parameters are fixed "by hand".
	An ablation study showing the sensitivity to hyper-parameters is provided in Section~\ref{sub:results} below.
	
	\paragraph{Metrics: mean-AP, AUC-PR and AUC-ROC.} 
	
	In order to evaluate the quality of the prediction of $y_i = [y_{i, 1}, \ldots, y_{i, B}]$ (when using ZiMM ED architecture or any other benchmark models) we use \emph{mean-AP}, defined as the average of the area under the precision-recall curve (AUC-PR), namely we compute the average over the buckets $b=1, \ldots, B$ of the AUC-PR for each bucket $b$, i.e., the AUC-PR of the prediction of $y_{i,b}$.
	Moreover, we report also the AUC-PR and the AUC-ROC (area under the ROC curve) for the binary classification problem $n_i > 0$ against $n_i = 0$.
	
	\paragraph{Train, validation and test sets.}
	
	All our experiments use the same random data splitting into 70\% of patients for training, 15\% of patients for validation and 15\% for testing. We checked the stationarity across the three splits of labels distribution and the main drugs, diagnoses and medical procedures.
	All the models are trained on a train set and we tune hyper-parameters using the validation set, while the test set is only used for the final evaluation.
	We report performances on both the validation and the test sets.

	\subsection{Baselines}
	\label{sub:baselines}
	
	The prediction performances of ZiMM-ED is compared with several baselines, involving several featuring strategies and different predictive models, both for binary prediction ($n_i > 0$ versus $n_i =0$) using the AUC-PR and AUC-ROC metrics and for the prediction of $y_i = [y_{i, 1}, \ldots, y_{i, B}]$ (the mean-AP score defined in Section \ref{sub:evaluation}). 
	Results are reported in Table~\ref{tab:results_baselines} below, where we observe that ZiMM ED improves all the considered baselines, in particular for the prediction of $y_i$, since the ZiMM model is dedicated to this task.
	We consider the following baseline models, more details are provided below:
	\begin{itemize} 
		\item LRl2: Logistic regression with $\ell_2$ penalization using the scikit-learn library~\cite{pedregosa2011scikit};
		\item GBDT: Gradient boosting using XGBoost~\cite{chen2015xgboost};
		\item MLP: Multilayer Perceptron model with 128 hidden units;
		\item Word2vec: we first train embeddings of all medical codes following~\cite{mikolov2013distributed}, and use these pre-trained embeddings in a MLP with 128 units;
		\item LSTM: a single embedding layer and one forward LSTM layer with 128 hidden units;
		\item Patient2Vec \cite{Zhang2018}.
	\end{itemize}
	We evaluated them using 3 types of input features described below.

	\paragraph{Static features (SF).} 
	
	This featuring is used for LRl2, GBDT and MLP models.
	Inputs correspond to aggregated counts of grouped medical codes over the entire observation period of a patient. 
	Number of occurrences of each code is then multiplied by the corresponding one-hot encoding. 
	Hence, the input is a $N$-dimensional vector representing a patient’s medical history. 
	One logistic regression (LRl2) and one gradient boosting decision tree (GBDT) is trained for each bucket $b$.
	This input is also used for the multi-layer perceptron (MLP).
	
	\paragraph{Dynamic features (DF).}
	
	This featuring is used for LRl2, GBDT and MLP models.
	We split the sequence into subsequences of 60 days, on which we compute the same features as with SF, but within each interval, in order to incorporate longitudinal information into LRl2, GBDT and MLP.
	
	\paragraph{Irregularly-spaced sequence (ISS).}
	
	This featuring is used for Word2vec, LSTM and Patient2Vec models.
	In this case, we consider the original patient sequence. 
	For both the Word2vec model and the LSTM one, the events are just gathered in a sequence in which the time-stamps or duration of the events are not used.
	In the Word2vec case, the codes embeddings are trained using Word2vec~\cite{mikolov2013distributed} and prediction is performed with a MLP with 128 hidden nodes.
	In the LSTM case, an embedding layer and one forward LSTM layer with 128 hidden units are trained in an end-to-end fashion and is applied to the sequence.
	The Patient2vec uses the same input features as the ZIMM ED architecture (i.e., sequence, including time-stamps and duration of events). 
	It has been only used for the binary classification problem (i.e., $n_i > 0$ versus $n_i = 0$).
	
	\medskip
	
	The predictive performance of all benchmarks and of ZiMM ED are presented in Table~\ref{tab:results_baselines}, where we report mean-AP scores on the test set, as well as AUC-PR and AUC-ROC scores for the binary classification problem (see Section \ref{sub:evaluation} for definition of these scores).
	According to this table, GBDT-based models on dynamic features (GBDT-DF) performs the best among all benchmark models, however the ZIMM ED architecture outperforms all the benchmark models both for the multi-output $y_i$ prediction and for the binary prediction.
	
	\begin{table}[htbp]
		\centering
		\small
		\begin{tabular}{l|lll} 
			\hline
			Model & mean-AP & AUC-ROC & AUC-PR\\ 
			\hline
			LRl2-SF     & 0.19 & 0.64 & 0.50\\
			GBDT-SF     & 0.24 & 0.67 & 0.56\\
			MLP-SF      & 0.18 & 0.64 & 0.49\\
			\hline
			LRl2-DF     & 0.21 & 0.65 & 0.53\\
			GBDT-DF     & 0.25 & 0.68 & 0.57 \\
			MLP-DF      & 0.19 & 0.65 & 0.50\\
			\hline
			Word2vec-ISS    & 0.20 & 0.65 & 0.53\\
			LSTM-ISS     & 0.21 & 0.67 & 0.54\\
			Patient2Vec & - & 0.68 & 0.55\\
			\hline
			ZiMM ED (best model) & \textbf{0.306} & \textbf{0.701} & \textbf{0.619} \\
			\hline
		\end{tabular}
		\caption{\small Predictive performances (on test data) of benchmark models and ZiMM ED architecture. ZIMM ED appears to perform the best among all models both for multi-output and binary prediction. mean-AP stands for average of the area under the precision-recall curve (AUC-PR) over buckets $b=1, \ldots, B$.}
		\label{tab:results_baselines}
	\end{table}
	
	Additionally, following suggestions in~\cite{mallik_graph_2020, bandyopadhyay_survey_2014}, we also report, in Figure~\ref{fig:CI_baselines} below, boxplots for the mean-AP score computed on 1000 bootstraped samples based on the test set, together with, in Table~\ref{tab:pvalues}, the $p$-values of pairwise Mann-Whitney U statistical tests for the null assumption that the mean-AP score of ZiMM ED and each baseline are equal. 
	We observe, both in Figure~\ref{fig:CI_baselines} and Table~\ref{tab:pvalues}, that ZiMM ED significantly improve the mean-AP scores of all the considered baselines: all the tests strongly reject the null hypothesis (the largest $p$-value is $< 10^{-6}$) and the boxplot of ZiMM ED exhibits a distribution which is significantly shifted towards higher mean-AP scores.
	
	\begin{figure}[h]
		\centering
		\includegraphics[width=0.95\textwidth]{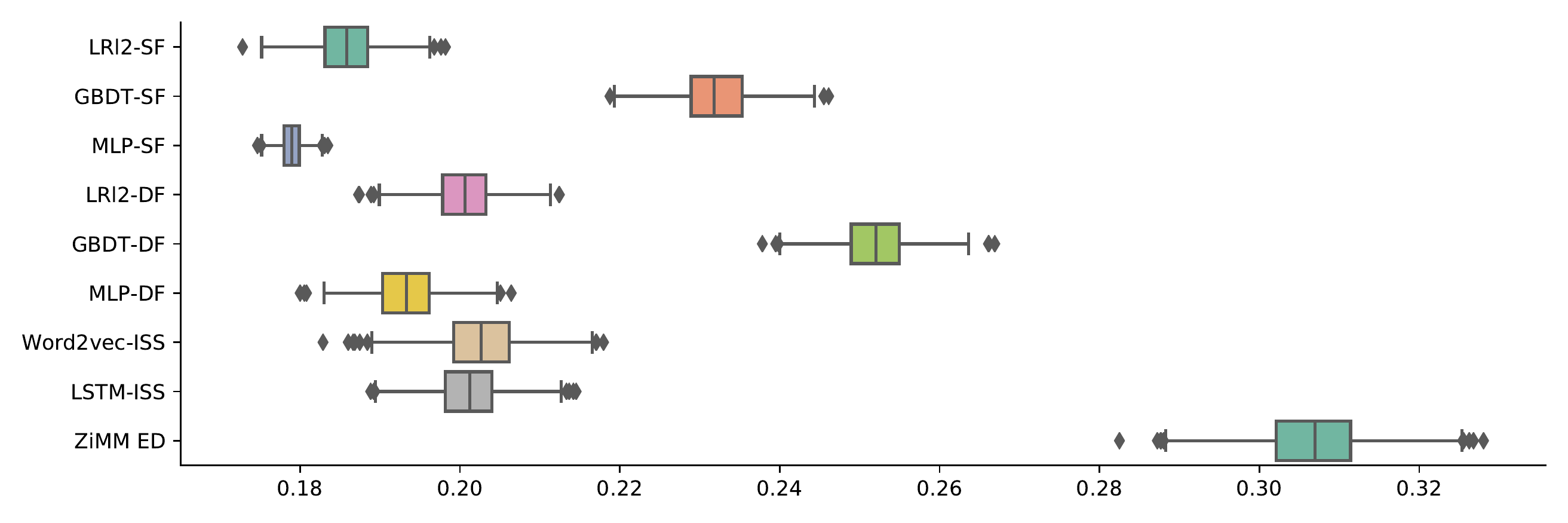}
		\caption{\small Boxplots of the mean-AP scores computed on $1000$ bootstraped samples based on the test set of all baseline models and ZiMM ED. The boxplot of ZiMM ED exhibits a distribution which is significantly shifted towards higher mean-AP scores.}
		\label{fig:CI_baselines}
	\end{figure}
	
\begin{table}
	\centering
	\small
	\begin{tabular}{cccccccc}\hline
		LRl2-SF & GBDT-SF &	MLP-SF & LRl2-DF & GBDT-DF & MLP-DF & Word2vec-ISS & LSTM-ISS \\ \hline
		0.0 &  5.32e-07 & 0.0 & 0.0 & 3.65e-13 & 0.0 & 0.0 & 0.0 \\ \hline
	\end{tabular}
	\caption{\small $p$-values of pairwise Mann-Whitney U statistical tests for the null assumptions that the mean-AP scores of ZiMM ED and each baseline are equal. All tests strongly reject this hypothesis. Note that zero $p$-values correspond to situations where the mean-AP score of the baseline was much smaller than the one of ZiMM ED for each boostrap sample.}
	\label{tab:pvalues}
\end{table}

	\subsection{Ablation study: performances of our model and variations around it}
	\label{sub:results}
	
	The performance reported in Table~\ref{tab:results_baselines} for the ZiMM ED architecture relies on a careful tuning of several hyper-parameters.
	This corresponds to the so-called \emph{ZiMM ED default} architecture, where the hyperparameters used are described in Table~\ref{tab:zimm-default} below.
	\begin{table}[h!]
		\small
		\centering
		\begin{tabular}{clc}
			\hline
			& Hyper-parameter & Value \\
			\hline
			\multirow{5}{*}{\rotatebox[origin=c]{90}{\parbox[c]{2.0cm}{\centering Data preprocessing}}}
			& Maximum \#days observed in patients' sequence before $T^i$ & 50 \\
			& Maximum \#medical events within the same day & 24 \\
			& Vocabulary size for medications coded with ATC  & 1036\\
			& Vocabulary size for diagnoses coded with ICD-10 & 1391 \\
			& Vocabulary size for medical procedures coded with CCAM & 1146 \\
			\hline
			\multirow{6}{*}{\rotatebox[origin=c]{90}{\parbox[c]{2.5cm}{\centering Embedding of medical codes and time}}}
			& Embedding dimension of medical codes & 64 \\
			& L2 penalization rate for medical codes embedding & 0.005 \\
			& Gaussian dropout rate & 0.3 \\
			& Embedding dimension for time horizon and hospitalization duration & 4 \\
			& L2 penalization rate for time embedding & 0.01 \\
			& Batch normalization epsilon & 1e-06 \\
			\hline
			\multirow{5}{*}{\rotatebox[origin=c]{90}{\parbox[c]{2.0cm}{\centering Embeddings aggregation}}}
			& Aggregation mode & self-attention \\
			& Number of heads & 3 \\
			& Weight drop-connect rate & 0.3 \\
			& Dropout rate & 0.2 \\
			& L2 penalization rate & 0.01 \\
			\hline
			\multirow{5}{*}{\rotatebox[origin=c]{90}{\parbox[c]{2.0cm}{\centering ZiMM Encoder}}}
			& Recurrent layer type & LSTM \\
			& \#hidden units per layer & 64 \\
			& \#layers & 1\\
			& Dropout rate & 0.3 \\
			& Recurrent dropout rate & 0.2\\
			\hline
			\multirow{6}{*}{\rotatebox[origin=c]{90}{\parbox[c]{2.5cm}{\centering ZiMM Decoder}}}
			& Recurrent layer type & GRU \\
			& \#hidden units per layer & 32 \\
			& \#RNN layers used in parallel for each value $n_i = b$ & 1\\
			& \#common RNN layers & 1\\
			& Gaussian dropout rate & 0.3 \\
			& Recurrent dropout rate & 0.2\\
			\hline
			\multirow{3}{*}{\rotatebox[origin=c]{90}{\parbox[c]{1.5cm}{\centering Training}}}
			& Optimizer type & Nadam \\
			& Learning rate & 0.001\\
			& Batch size & 256\\[5pt]
			\hline
		\end{tabular}
		\caption{\small ZiMM-ED default parameters.}
		\label{tab:zimm-default}
	\end{table}
	
	The hyper-parameters described in Table~\ref{tab:zimm-default} have been selected using an extensive random search for the best mean-AP metric on the validation set.
	In Figure~\ref{fig:random_search} below, we illustrate the value of this metric ($y$-axis) for some models that were evaluated during the random search.
	Each point corresponds to a single model with fixed hyper-parameters, and the set of models is the same on all four displays of Figure~\ref{fig:random_search}.
	In each of these displays we ``align'' all the models that share a specific hyper-parameter, namely, respectively from left to right: the number of heads used in the self-attention layer, the number of hidden units used in the recurrent layer of the ZiMM Encoder, the number of stacked recurrent layers in ZiMM Encoder and finally the number of hidden units used in the ZiMM Decoder.
	The red point on all four Figures corresponds to the best model overall, that leads to the ZiMM ED default architecture described in Table~\ref{tab:zimm-default}.
	Finally, note that the random search included many other hyper-parameters, such as several learning-rate scheduling strategies, dropout rates, types of dropout regularization including input, output, embedding, cell-state and multiplicative Gaussian, that we do not report for the sake of conciseness.
	
	\begin{figure}[h]
		\centering
		\includegraphics[width=0.35\textwidth]{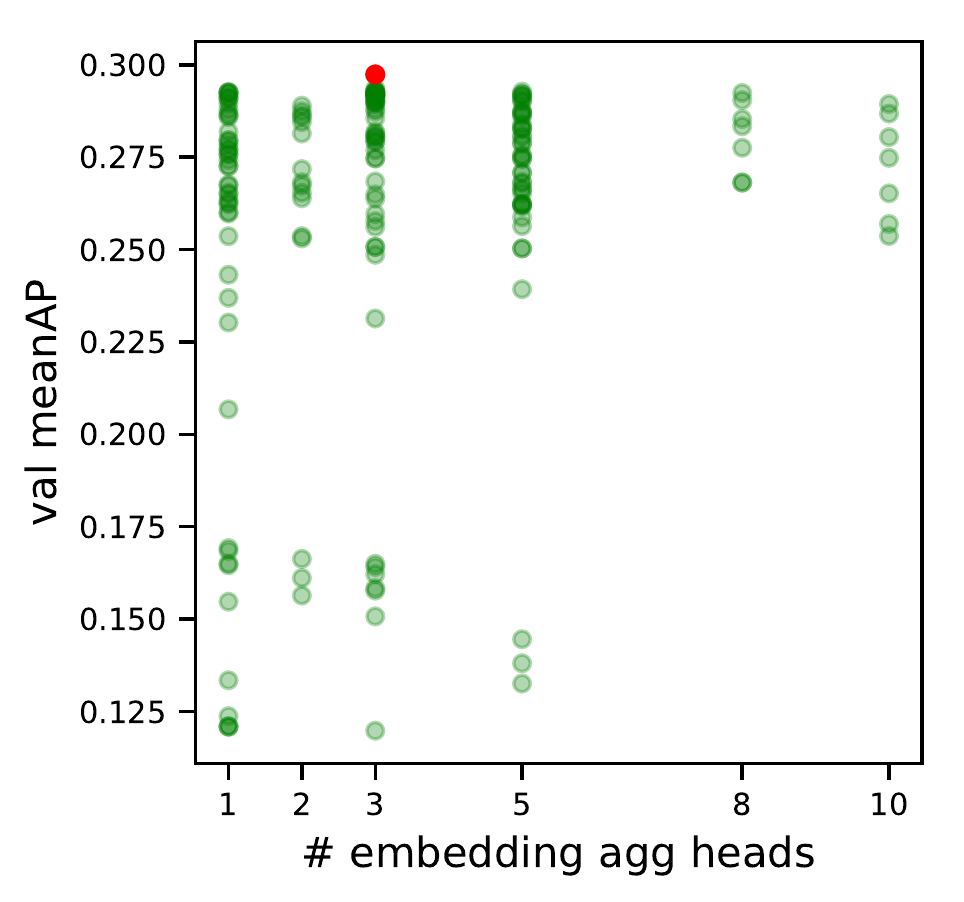}
		\includegraphics[width=0.35\textwidth]{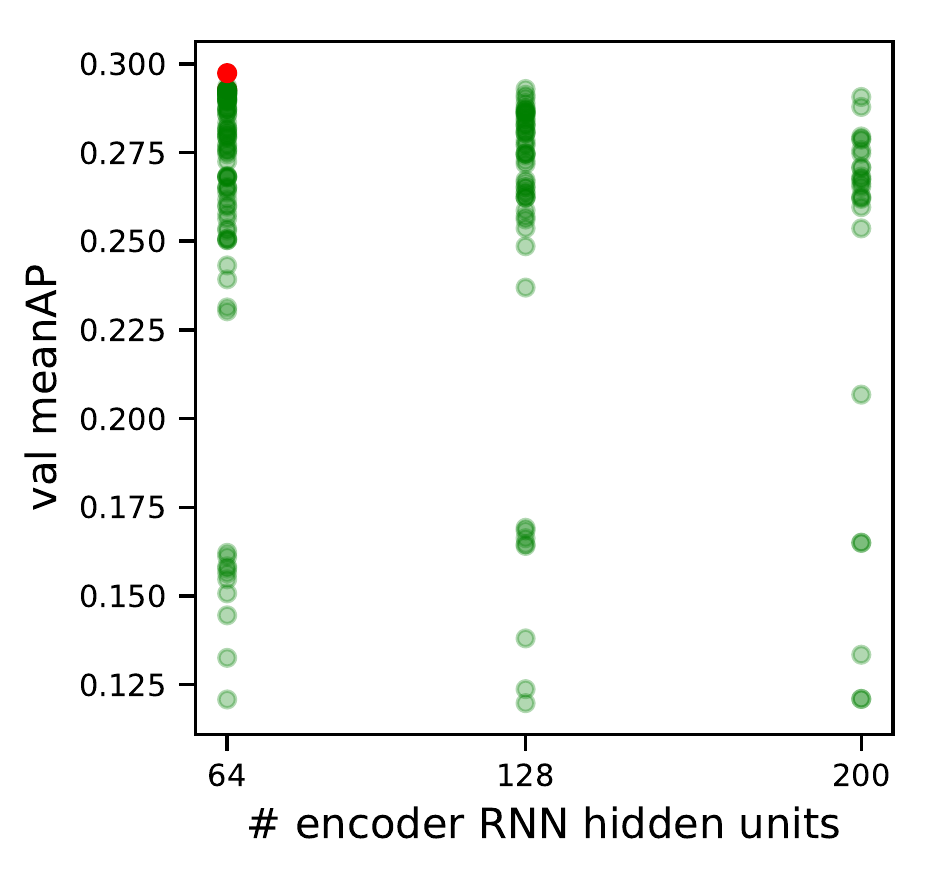}
		\includegraphics[width=0.35\textwidth]{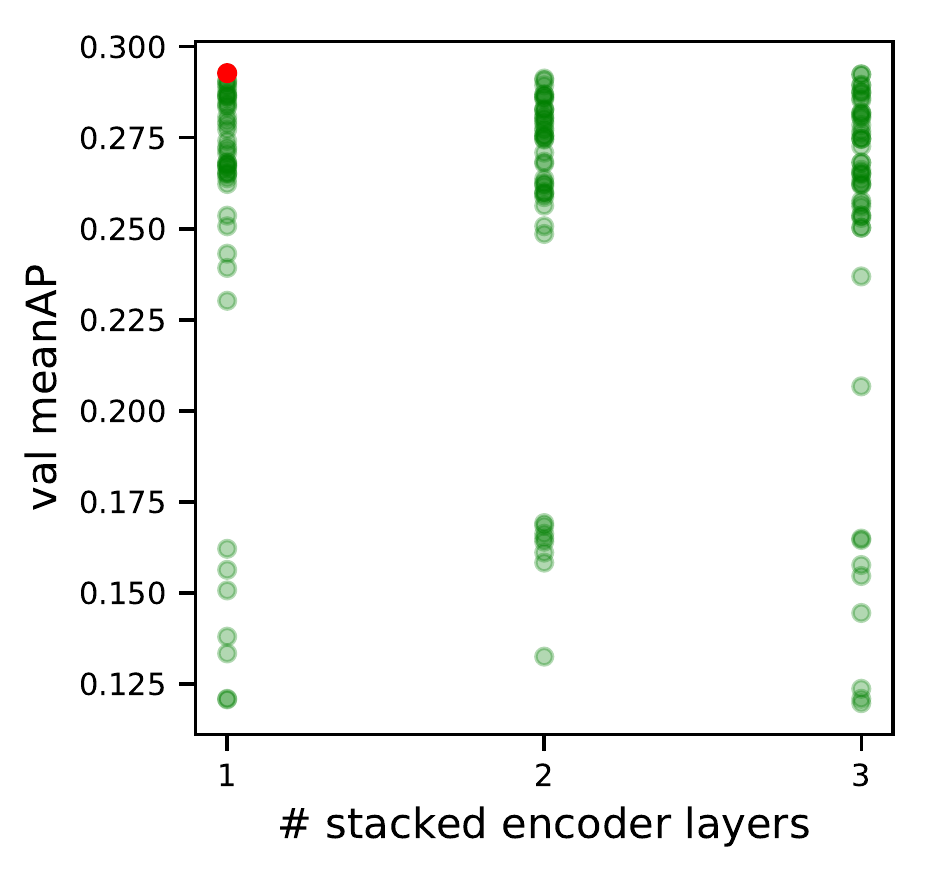}
		\includegraphics[width=0.35\textwidth]{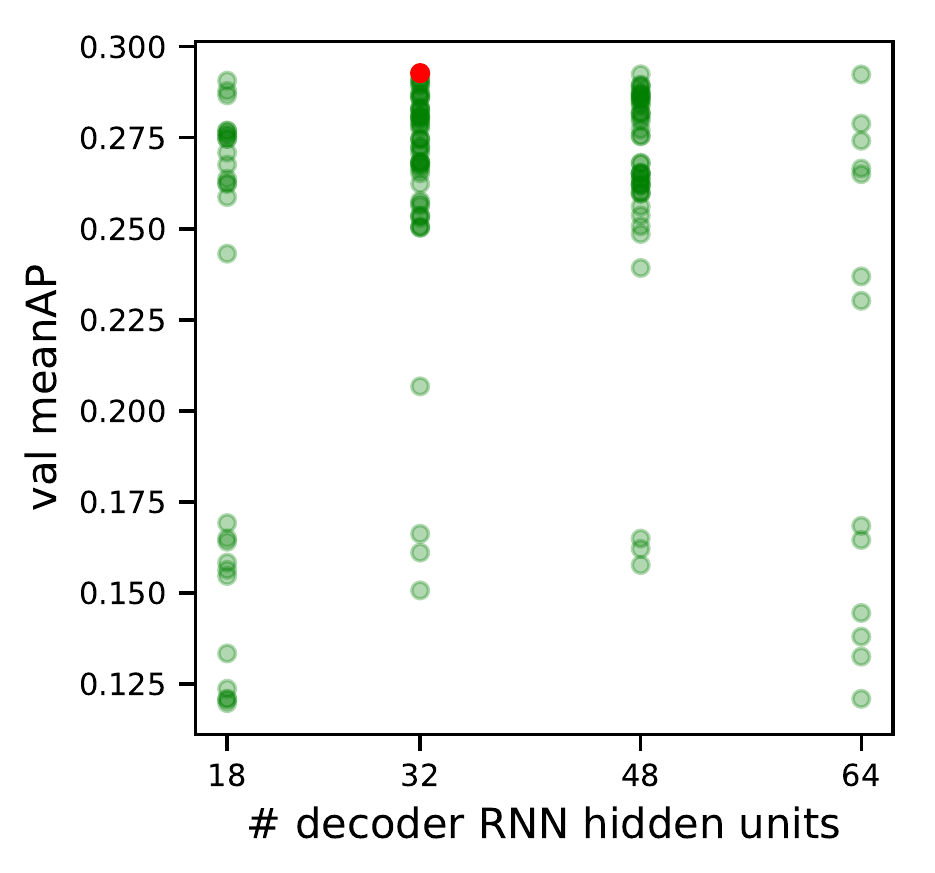}
		\caption{\small Validation mean-AP ($y$-axis) of some models evaluated during random search, where each point corresponds to a single model. The set of models is the same on all four displays and the best one (ZiMM ED default) is indicated by a red point. Each display aligns all the models that share a specific hyper-parameter, being, respectively from left to right, the number of heads used in the self-attention layer, the number of hidden units used in the recurrent layer of the ZiMM Encoder, the number of stacked recurrent layers in ZiMM Encoder and finally the number of hidden units used in the ZiMM Decoder.}
		\label{fig:random_search}
	\end{figure}

	The remaining of this section proposes an ablation study: we modify some hyper-parameters or change some components of ZiMM ED default, and report the impacts on performances in Table~\ref{tab:model_variations} below, where the first line reports the performances of ZiMM ED default.
	A discussion around these ablations is given below.
	
	\paragraph{Data preprocessing.}
	
	The first part (PP) of Table~\ref{tab:model_variations} shows results obtained with variations of the maximum sequence lengths and the classification system used for drugs encoding.
	The ZiMM ED default uses ATC encoding instead of the raw CIP-13 encoding, which allows to reduce the vocabulary size from 10 664 for CIP-13 to 1 105 for ATC.
	As observed in Table~\ref{tab:model_variations}, using CIP-13 (which is a much larger vocabulary for drugs) instead of ATC actually hurts strongly the performances.
	We observe also that considering longer sequences (100 days) instead of what we do in ZiMM ED default (50 days) deteriorates the performances as well.
	
	\paragraph{Embedding of medical codes and time.}
	
	Variations around the way embeddings are produced for medical codes and time are reported in the second part (Embedding) of Table~\ref{tab:model_variations}.
	We observe that increasing the embedding dimensions does not offer performance increase (using $E_\text{dim}=128$ instead of the default 
	$E_\text{dim} = 64$), as well as using the same embedding space for all medical codes. 
	In the ZiMM ED default model, tokenized durations and distance to $T^i$ (see Section~\ref{sub:zimm-encoder} above) are embedded with trainable embedding vectors. 
	We observe in Table~\ref{tab:model_variations} that without these time embeddings, the model performance deteriorates. 
	Also, we find that using trigonometric functions for ``positional encoding'' as proposed in~\cite{Vaswani2017} instead of learned embeddings does not help in our setting.

	\paragraph{Embeddings aggregation.}
	
	Results obtained through different embeddings aggregation techniques are presented in the third part (Aggregation) of Table~\ref{tab:model_variations}.
	We applied penalization and dropout on attention weights as proposed in~\cite{Lin2019}, as well as DropConnect~\cite{Wan2013} for regularizing weights in large fully-connected layers.
	As explained in~\cite{Vaswani2017}, multi-head attention allows the model to jointly attend information from different representation subspaces at different positions, however in our setting, increasing SA $N_{\text{heads}}$ (self-attention number of heads) in the self-attention layer does not clearly improve the performance.
	We tested $N_{\text{heads}} = 8$, as reported in~\cite{Luo2019}, but the best performance in our case was achieved with $N_{\text{heads}} = 3$.
	
	\paragraph{ZiMM Encoder.}
	
	We report in the fourth part (Encoder) of Table~\ref{tab:model_variations} the results of extensive experiments performed in order to identify the best combination of hyper-parameters for both recurrent and convolutional layers (CNN): number of hidden units, types of recurrent layers and different dropout rates used in the recurrent layers. 
	Only the best results for each type of layer are reported in Table~\ref{tab:model_variations}.
	We observe that GRU layers perform generally as well as LSTM layers. 
	Increasing the number of hidden units in recurrent layers to $N_\text{units} = 128$ instead of the default choice $N_\text{units} = 64$ does not significantly improve performance, as well as using two stacked LSTM layers $N_{\text{layers}}=2$ instead of the default $N_{\text{layers}}=1$.
	We observed also that in our setting, CNN layers badly under-perform compared to the other types of layers.
	Finally, we tested ``Multi-head transformer'' which corresponds to the transformer encoder~\cite{Vaswani2017} which takes as input all patient sequence of events. Clinical codes are embedded in the same way as for ZiMM model, which is then passed to the encoder block with a 4-head self-attention mechanism through the entire patient sequence.
	
	\begin{table}[h!]
		\small
		\centering
		\begin{tabular}{c|l|l|l|lll} 
			\hline
			& \multirow{2}{*}{Architecture modification} & \# params & val set   & \multicolumn{3}{c}{test set} \\
			& & $ \times 10^{5}$  & mean-AP   & mean-AP & AUC-ROC & AUC-PR \\
			\hline
			& \emph{ZiMM ED default}  & 3.5 & 0.292 & 0.304 & 0.704 & 0.619 \\ 
			\hline
			\multirow{2}{*}{\rotatebox[origin=c]{90}{\parbox[c]{0.5cm}{\centering PP*}}} 
			& CIP-13 drugs encoding   & 7.1 & 0.279 & 0.286 & 0.690 & 0.604 \\
			& 100-days sequence        & 3.5 & 0.291 & 0.300 & 0.701 & 0.616 \\
			\hline
			\multirow{4}{*}{\rotatebox[origin=c]{90}{\parbox[c]{1.2cm}{\centering{\scriptsize Embedding}}}} 
			& Common embedding space  & 13.9 & 0.292 & 0.298 & 0.698 & 0.613 \\
			& Without $\Delta t$ embedding & 3.5 & 0.293 & 0.300 & 0.702 & 0.617 \\
			& Positional encoding   & 3.5 & 0.292 & 0.302 & 0.702 & 0.620 \\
			& $E_{\text{dim}} = 128$    & 7.4 & 0.291 & 0.300 & 0.701 & 0.615 \\
			\hline
			\multirow{5}{*}{\rotatebox[origin=c]{90}{\parbox[c]{1cm}{\centering {\scriptsize Aggregation}}}}
			& SA $N_{\text{heads}} = 1$  & 3.2 & 0.293 & 0.303 & 0.701 & 0.619 \\
			& SA $N_{\text{heads}} = 5$  & 3.7 & 0.290 & 0.300 & 0.704 & 0.620 \\
			& SA $N_{\text{heads}} = 8$  & 4.1 & 0.292 & 0.300 & 0.701 & 0.616 \\
			& mean              & 3.1 & 0.287 & 0.297 & 0.696 & 0.612 \\
			\hline
			\multirow{6}{*}{\rotatebox[origin=c]{90}{\parbox[c]{1cm}{\centering {Encoder}}}}
			& $N_{\text{units}} = 128$  & 4.6 & 0.293 & 0.301 & 0.701 & 0.618 \\
			& $N_{\text{layers}} = 2$   & 3.8 & 0.290 & 0.306 & 0.701 & 0.619 \\
			& bi-LSTM                   & 4.2 & 0.290 & 0.300 & 0.700 & 0.617 \\
			& GRU                       & 3.3 & 0.294 & 0.300 & 0.702 & 0.614 \\
			& Conv1D                    & 3.2 & 0.231 & 0.234 & 0.649 & 0.536 \\
			& Multi-head transformer    & 6.2 & 0.279 & 0.287 & 0.694 & 0.607 \\
			\hline
			\multirow{5}{*}{\rotatebox[origin=c]{90}{\parbox[c]{1cm}{\centering {Decoder}}}}
			& FC layer    & 3.3 & 0.290 & 0.300 & 0.691 & 0.617\\
			& LSTM        & 3.5 & 0.291 & 0.301 & 0.703 & 0.619 \\
			& basic RNN   & 3.4 & 0.292 & 0.301 & 0.701 & 0.619 \\
			& $N_{\text{units}} = 18$ & 3.4  & 0.292 & 0.298 & 0.702 & 0.619  \\
			& $N_{\text{layers}} = 2$ & 3.5 & 0.288 & 0.297 & 0.699 & 0.615 \\
			\hline
		\end{tabular}
		\caption{\small Performances of some variations around the ZiMM ED default architecture, for which all hyper-parameters are given in Table~\ref{tab:zimm-default} above. *PP stands for ``preprocessing''. SA $N_{\text{heads}}$ stands for number of self-attention heads; bi-LSTM stands for bidirectional LSTM; mean-AP stands for average of the area under the precision-recall curve (AUC-PR) over the buckets $b=1, \ldots, B$; AUC-ROC stands for area rnder the Receiver operating characteristics curve.}
		\label{tab:model_variations}
	\end{table}
	
	\paragraph{ZiMM Decoder.}
	
	Finally, we consider some variations around the ZiMM Decoder, and report the results in the fifth part (Decoder) of Table~\ref{tab:model_variations}.
	While the ZiMM Decoder described in Section~\ref{sub:zimm-decoder} uses a single feed-forward layer to predict the mixture probabilities, a GRU layer to learn hidden states and GRU layers to predict the parameters of each multinomial distributions, one could use instead fully-connected layers to predict all the parameters of the ZiMM distribution, or alternatively simple RNN layers or LSTM layers.
	The fully-connected layers deteriorates the most the performance, while replacing the GRU layers by RNN or LSTM layers only deteriorates it mildly.
	We also report the performance obtained with two stacked GRU $N_{\text{layers}} = 2$ instead of the default one $N_{\text{layers}} = 1$ to produce the hidden states (see Equation~\eqref{eq:decoder_hidden}) and with a smaller hidden size $N_{\text{units}} = 18$ instead of  $N_{\text{units}} = 32$.

	\subsection{Visualization of the embeddings produced for diagnoses and drugs codes}

\begin{figure}[htbp]
		\centering
		\includegraphics[width=0.5\textwidth]{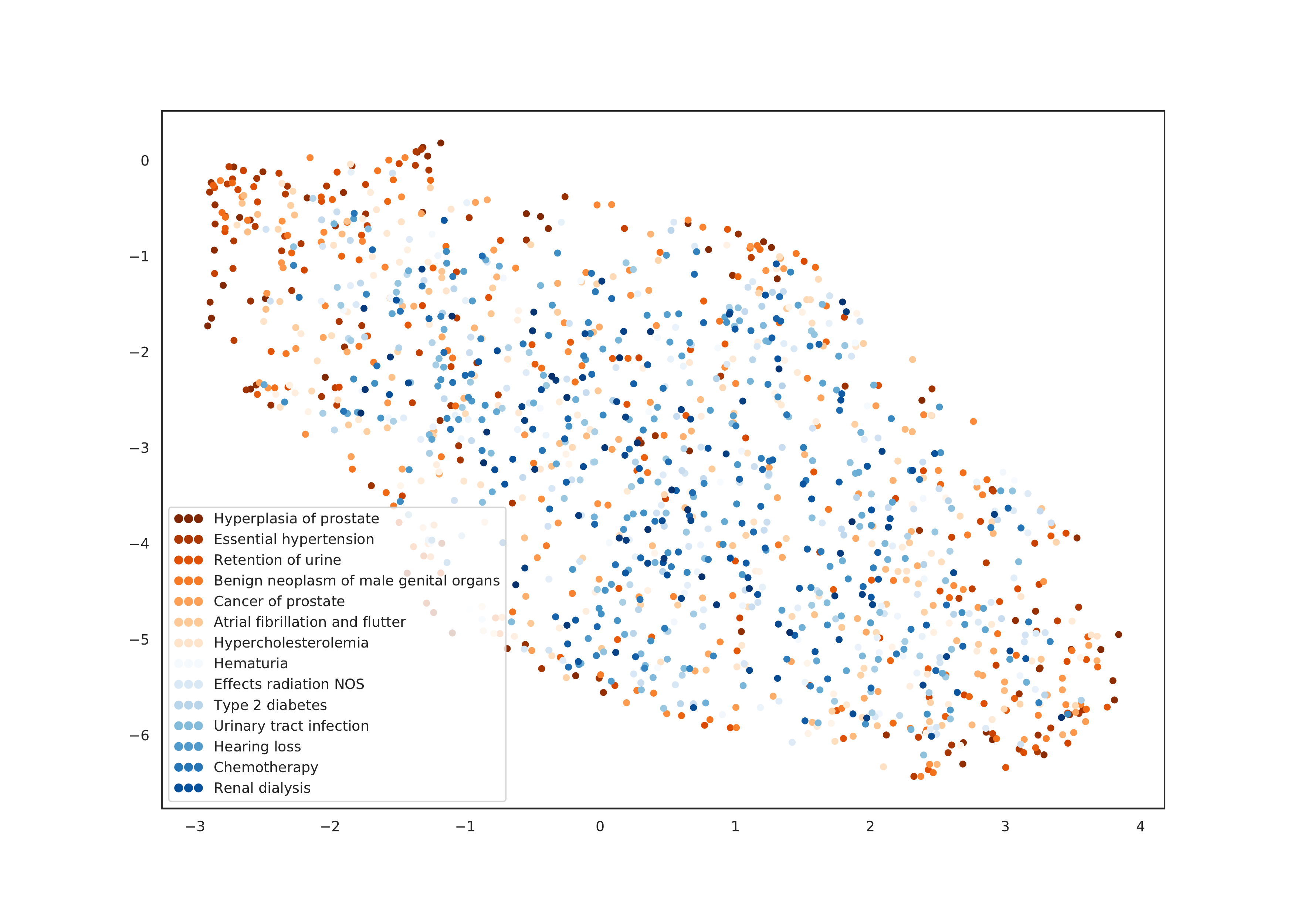}
		\includegraphics[width=0.469\textwidth]{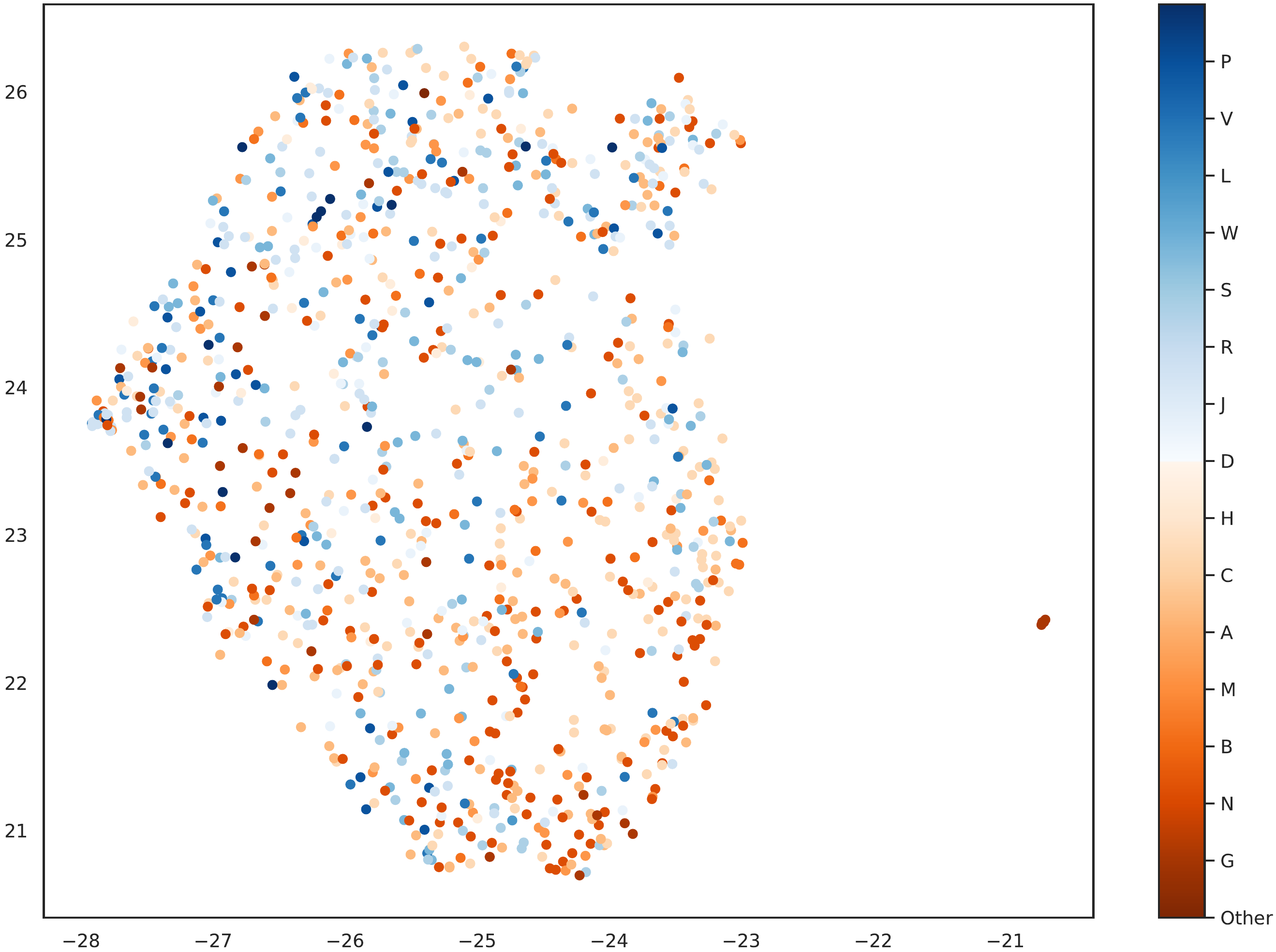}
		\caption{\small UMAP projections of the embeddings of medical codes. \emph{Left-hand side.} Each point is the projection of the embedding vector of an ICD code, colored with its corresponding PheWAS phenotypes. Note that many points share the same color since many ICD10 codes have the same PheWAS. The legend provides only the most common diagnoses. \emph{Right-hand side.} UMAP projections of the drugs embeddings, colored by the first level of the ATC classification (one letter).}
		\label{fig:embedding_umap}
	\end{figure}

	We explore the embeddings produced for diagnoses and drugs as a by-product of the ZiMM ED architecture (see Section~\ref{sub:embeddings} about the embeddings of medical codes).
	We use UMAP~\cite{mcinnes_umap_2018} in order to reduce the dimension from 64 to 2. 
	The resulted projections are shown in Figure~\ref{fig:embedding_umap}. 
	The UMAP algorithm requires four hyper-parameters: the number of neighbors to consider when approximating the local metric (n-neighbors),
	the desired separation between close points in the embedding space (min-dist), the number of training epochs (n-epochs) and finally the projection dimension ($d)$. 
	We fixed \verb|d=2| and \verb|n-epochs = 500|. ICD10 disease codes are mapped in PheWAS phenotypes~\cite{Diogo2018} and drugs codes are mapped to the first level of the ATC classification (main anatomical group consisting of a single letter), leading to the color scheme used in Figure~\ref{fig:embedding_umap}.
	
	On the left-hand side of Figure~\ref{fig:embedding_umap}, we can observe the UMAP projections of the embedding vectors of each ICD10 code, colored by the corresponding PheWAS phenotype.
	The hyper-parameters used here for UMAP are \verb|n-neighbors=20| and \verb|min-dist=0.1|.
	Many points share the same color since many ICD10 codes have the same PheWAS.
	We observe here that diagnoses that are known to co-occur and belong to the same clinical groups are projected close to each other.
	The red-orange-highlighted points are the phenotypes the most related to the urogenital system such as \emph{Hyperplasia of prostate}, \emph{Retention of urine}, \emph{Benign neoplasm of male genital organs}, and \emph{Cancer of prostate}. 
	We can also observe a visually strong association of these phenotypes with \emph{Essential hypertension}.
	This is confirmed by~\cite{michel_association_2004}, where the association of hypertension and symptoms of benign prostatic hyperplasia is well-studied and understood.
	Other clinical concepts that are projected close to each other are \emph{Hearing loss} and \emph{Chemotherapy}, \emph{Urinary tract infection} and \emph{Hematuria} (blue-highlighted points), which is also confirmed by~\cite{taborelli_increased_2019, lin_association_2015}, where
	it is shown that renal dialysis is associated with a higher risk of cancer. 
	
	On the right-hand side of Figure~\ref{fig:embedding_umap}, we observe the UMAP projections of drugs embeddings, colored by the first level of the ATC classification (first letter).
	The UMAP hyper-parameters used here are \verb|n-neighbors=50| and \verb|min-dist=0.2|.
	The dark blue points correspond to \textit{Various} ATC class (V), that mainly correspond to contrast agents for MRI.
	We observe that these points do not form any cluster but are dispersed over the whole space.
	This can be explained by the fact that the considered cohort contain only men with BPH (see Section~\ref{sub:cohort}), that often have an MRI of the prostate. 
	Furthermore, we observe that the drugs from the class \textit{Genito-urinary system and sex hormones} (G), that include drugs for urination problems, form one cluster with other age-related medications from the class \textit{Nervous system} (N), namely analgesics, anti-thrombotics, psycholeptics, psychotropic, and anti-parkinsonian drugs.

	\section{Conclusion and future works}
	\label{sec:conclusion}

	In this work, we propose ZiMM Encoder Decoder (ZiMM ED), an end-to-end deep learning model trained against the negative log-likelihood of the new ZiMM (Zero-inflated Mixture of Multinomial) model, for the modeling of long-term and blurry relapses. 
	This deep learning model is trained on a cohort based on a large electronic health record database, that contains only claims and no clinical data, from the whole French population.
	
	We show that our model improves the performances of a large number of baselines, including the state-of-the-art, for the considered predictive task.
	ZiMM ED allows to represent the full health pathways of patients, using all available information, with minimal preprocessing.
	Therefore, this end-to-end architecture can be used for other tasks as well, through transfer learning or fine tuning of the model.
	Future works will consider a multi-task version of ZiMM ED (several types of relapses, or other types of events), and other improvements using alternative architectures for attention modeling~\cite{kitaev_reformer_2020, dai_transformer-xl_2019, lan_albert_2020, wu_combiner_2019}.
	
	The considered dataset and the introduced methodology was developped for non-clinical claims data, which is fundamentally different from clinical datasets. A future effort will be done with the aim of adding a clinical and genomic data to th cohort. In this case a lot of additions analysis can be done (\cite{bhadra_identification_2019}).
	
	This work addresses the problem of predicting the blurry relapses of the TURP surgery, which is the first step towards an evidence-based approach using machine-learning to help the clinical decision.
	The next step is to exploit these predictions to help to decide the timing of the surgery: given the current health pathway of the patient, what is his probability of a relapse, so that it can help the clinician to decide when to perform the surgery, or to consider alternative treatments.

	\subsection*{Acknowledgments}
	
	This research benefited from the National Health Insurance Fund (CNAM) partnership with Data Science initiative at École polytechnique. We thank Maryan Morel, Dinh Phong Nguyen, Youcef Sebiat, Dian Sun, researchers and engineers working on this project. 
	Yiyang Yu was supported by grants from Région Ile-de-France. 
	Anastasiia Kabeshova was supported by CNAM-Polytechnique research partnership.
	This research is also supported by the Agence Nationale de la Recherche as part of the ``Investissements d'avenir'' program (reference ANR-19-P3IA-0001; PRAIRIE 3IA Institute).
	Finally, we wish to thank Pr. Eric Vicaut, for his help in the understanding and the usage of the database.


\small
	
\begin{thebibliography}{73}
\providecommand{\natexlab}[1]{#1}
\providecommand{\url}[1]{\texttt{#1}}
\expandafter\ifx\csname urlstyle\endcsname\relax
  \providecommand{\doi}[1]{doi: #1}\else
  \providecommand{\doi}{doi: \begingroup \urlstyle{rm}\Url}\fi

\bibitem[Abadi et~al.(2015)Abadi, Agarwal, Barham,
  et~al.]{tensorflow2015-whitepaper}
M.~A. Abadi, A.~Agarwal, P.~Barham, et~al.
\newblock {TensorFlow}: Large-scale machine learning on heterogeneous systems,
  2015.
\newblock URL \url{https://www.tensorflow.org/}.
\newblock Software available from tensorflow.org.

\bibitem[Atramont et~al.(2018)Atramont, Bonnet-Zamponi, Bourdel-Marchasson,
  Tangre, Fagot-Campagna, and Tuppin]{Atramont2018}
A.~Atramont, D.~Bonnet-Zamponi, I.~Bourdel-Marchasson, I.~Tangre,
  A.~Fagot-Campagna, and P.~Tuppin.
\newblock {Health status and drug use 1 year before and 1 year after skilled
  nursing home admission during the first quarter of 2013 in France: a study
  based on the French National Health Insurance Information System.}
\newblock \emph{European journal of clinical pharmacology}, 74\penalty0
  (1):\penalty0 109--118, jan 2018.
\newblock ISSN 1432-1041.
\newblock \doi{10.1007/s00228-017-2343-y}.
\newblock URL \url{http://www.ncbi.nlm.nih.gov/pubmed/28975381}.

\bibitem[Ba et~al.(2016)Ba, Kiros, and Hinton]{Ba2016}
J.~L. Ba, J.~R. Kiros, and G.~E. Hinton.
\newblock {Layer Normalization}.
\newblock \emph{NIPS 2016 Deep Learning Symposium}, jul 2016.
\newblock URL \url{http://arxiv.org/abs/1607.06450}.

\bibitem[Bacry et~al.(2019)Bacry, Ga{\"{i}}ffas, Leroy, Morel, Nguyen, Sebiat,
  and Sun]{scalpel2019}
E.~Bacry, S.~Ga{\"{i}}ffas, F.~Leroy, M.~Morel, D.~P. Nguyen, Y.~Sebiat, and
  D.~Sun.
\newblock {SCALPEL3: a scalable open-source library for healthcare claims
  databases}.
\newblock pages 1--14, 2019.
\newblock URL \url{http://arxiv.org/abs/1910.07045}.

\bibitem[Bajor et~al.(2018)Bajor, Mesa, Osterman, and Lasko]{Bajor2018}
J.~M. Bajor, D.~A. Mesa, T.~J. Osterman, and T.~A. Lasko.
\newblock {Embedding Complexity In the Data Representation Instead of In the
  Model: A Case Study Using Heterogeneous Medical Data}.
\newblock feb 2018.
\newblock URL \url{http://arxiv.org/abs/1802.04233}.

\bibitem[Bandyopadhyay et~al.(2014)Bandyopadhyay, Mallik, and
  Mukhopadhyay]{bandyopadhyay_survey_2014}
S.~Bandyopadhyay, S.~Mallik, and A.~Mukhopadhyay.
\newblock A {{Survey}} and {{Comparative Study}} of {{Statistical Tests}} for
  {{Identifying Differential Expression}} from {{Microarray Data}}.
\newblock \emph{IEEE/ACM Transactions on Computational Biology and
  Bioinformatics}, 11\penalty0 (1):\penalty0 95--115, Jan. 2014.
\newblock ISSN 1545-5963.
\newblock \doi{10.1109/TCBB.2013.147}.

\bibitem[Baytas et~al.(2017)Baytas, Xiao, Zhang, Wang, Jain, and
  Zhou]{Baytas2017PatientSV}
I.~M. Baytas, C.~Xiao, X.~Zhang, F.~Wang, A.~K. Jain, and J.~Zhou.
\newblock Patient subtyping via time-aware lstm networks.
\newblock In \emph{KDD}, 2017.

\bibitem[Beam et~al.(2018)Beam, Kompa, Schmaltz, Fried, Weber, Palmer, Shi,
  Cai, and Kohane]{Beam2018}
A.~L. Beam, B.~Kompa, A.~Schmaltz, I.~Fried, G.~Weber, N.~P. Palmer, X.~Shi,
  T.~Cai, and I.~S. Kohane.
\newblock {Clinical Concept Embeddings Learned from Massive Sources of
  Multimodal Medical Data}.
\newblock apr 2018.
\newblock URL \url{http://arxiv.org/abs/1804.01486}.

\bibitem[Beaulieu-Jones and Greene(2016)]{Beaulieu-Jones2016}
B.~K. Beaulieu-Jones and C.~S. Greene.
\newblock {Semi-supervised learning of the electronic health record for
  phenotype stratification}.
\newblock \emph{Journal of Biomedical Informatics}, 64:\penalty0 168--178, may
  2016.
\newblock ISSN 15320464.
\newblock \doi{10.1016/j.jbi.2016.10.007}.

\bibitem[Bender and Sartipi(2013)]{Bender:2013}
D.~Bender and K.~Sartipi.
\newblock {HL7 FHIR}: An agile and {RESTful} approach to healthcare information
  exchange.
\newblock In \emph{Proceedings of CBMS 2013 - 26th IEEE International Symposium
  on Computer-Based Medical Systems}, pages 326--331. IEEE, jun 2013.

\bibitem[Bezin et~al.(2017)Bezin, Duong, Lassalle, Droz, Pariente, Blin, and
  Moore]{Bezin2017}
J.~Bezin, M.~Duong, R.~Lassalle, C.~Droz, A.~Pariente, P.~Blin, and N.~Moore.
\newblock {The national healthcare system claims databases in France, SNIIRAM
  and EGB: Powerful tools for pharmacoepidemiology.}
\newblock \emph{Pharmacoepidemiology and drug safety}, 26\penalty0
  (8):\penalty0 954--962, aug 2017.
\newblock ISSN 1099-1557.
\newblock \doi{10.1002/pds.4233}.
\newblock URL \url{http://www.ncbi.nlm.nih.gov/pubmed/28544284}.

\bibitem[Bhadra et~al.(2019)Bhadra, Mallik, and
  Bandyopadhyay]{bhadra_identification_2019}
T.~Bhadra, S.~Mallik, and S.~Bandyopadhyay.
\newblock Identification of {{Multiview Gene Modules Using Mutual
  Information}}-{{Based Hypograph Mining}}.
\newblock \emph{IEEE Transactions on Systems, Man, and Cybernetics: Systems},
  49\penalty0 (6):\penalty0 1119--1130, June 2019.
\newblock ISSN 2168-2216, 2168-2232.
\newblock \doi{10.1109/TSMC.2017.2726553}.

\bibitem[Chen et~al.(2015)Chen, He, Benesty, Khotilovich, and
  Tang]{chen2015xgboost}
T.~Chen, T.~He, M.~Benesty, V.~Khotilovich, and Y.~Tang.
\newblock Xgboost: extreme gradient boosting.
\newblock \emph{R package version 0.4-2}, pages 1--4, 2015.

\bibitem[Cheng et~al.(2016)Cheng, Wang, Zhang, and Hu]{Cheng2016}
Y.~Cheng, F.~Wang, P.~Zhang, and J.~Hu.
\newblock {Risk Prediction with Electronic Health Records: A Deep Learning
  Approach}.
\newblock In \emph{Proceedings of the 2016 SIAM International Conference on
  Data Mining}, pages 432--440, Philadelphia, PA, jun 2016. Society for
  Industrial and Applied Mathematics.
\newblock ISBN 978-1-61197-434-8.
\newblock \doi{10.1137/1.9781611974348.49}.
\newblock URL \url{https://epubs.siam.org/doi/10.1137/1.9781611974348.49}.

\bibitem[Cho et~al.(2014)Cho, Van~Merri{\"e}nboer, Gulcehre, Bahdanau,
  Bougares, Schwenk, and Bengio]{cho2014learning}
K.~Cho, B.~Van~Merri{\"e}nboer, C.~Gulcehre, D.~Bahdanau, F.~Bougares,
  H.~Schwenk, and Y.~Bengio.
\newblock Learning phrase representations using rnn encoder-decoder for
  statistical machine translation.
\newblock \emph{arXiv preprint arXiv:1406.1078}, 2014.

\bibitem[Choi et~al.(2016{\natexlab{a}})Choi, Bahadori, Schuetz, Stewart, and
  Sun]{Choi2015}
E.~Choi, M.~T. Bahadori, A.~Schuetz, W.~F. Stewart, and J.~Sun.
\newblock {Doctor AI: Predicting Clinical Events via Recurrent Neural
  Networks.}
\newblock \emph{JMLR workshop and conference proceedings}, 56:\penalty0
  301--318, nov 2016{\natexlab{a}}.
\newblock ISSN 1938-7288.
\newblock URL \url{http://arxiv.org/abs/1511.05942
  http://www.ncbi.nlm.nih.gov/pubmed/28286600{\%}0Ahttp://www.pubmedcentral.nih.gov/articlerender.fcgi?artid=PMC5341604}.

\bibitem[Choi et~al.(2016{\natexlab{b}})Choi, Bahadori, Searles, Coffey,
  Thompson, Bost, Tejedor-Sojo, and Sun]{Choi2016}
E.~Choi, M.~T. Bahadori, E.~Searles, C.~Coffey, M.~Thompson, J.~Bost,
  J.~Tejedor-Sojo, and J.~Sun.
\newblock {Multi-layer representation learning for medical concepts}.
\newblock \emph{Proceedings of the ACM SIGKDD International Conference on
  Knowledge Discovery and Data Mining}, 13-17-August-2016:\penalty0 1495--1504,
  feb 2016{\natexlab{b}}.
\newblock \doi{10.1145/2939672.2939823}.
\newblock URL \url{http://arxiv.org/abs/1602.05568}.

\bibitem[Choi et~al.(2016{\natexlab{c}})Choi, Bahadori, Sun, Kulas, Schuetz,
  Stewart, Sun, Kulas, Schuetz, Stewart, and Sun]{Choi2016retain}
E.~Choi, M.~T. Bahadori, J.~Sun, J.~Kulas, A.~Schuetz, W.~F. Stewart, J.~Sun,
  J.~Kulas, A.~Schuetz, W.~F. Stewart, and J.~Sun.
\newblock {RETAIN: Interpretable Predictive Model in Healthcare using Reverse
  Time Attention Mechanism}.
\newblock In \emph{Advances in Neural Information Processing Systems}, number
  Nips, pages 3504--3512, 2016{\natexlab{c}}.

\bibitem[Choi et~al.(2017)Choi, Bahadori, Song, Stewart, and Sun]{Choi2016gram}
E.~Choi, M.~T. Bahadori, L.~Song, W.~F. Stewart, and J.~Sun.
\newblock {GRAM: Graph-based attention model for healthcare representation
  learning}.
\newblock \emph{Proceedings of the ACM SIGKDD International Conference on
  Knowledge Discovery and Data Mining}, Part F129685:\penalty0 787--795, nov
  2017.
\newblock \doi{10.1145/3097983.3098126}.
\newblock URL \url{http://arxiv.org/abs/1611.07012}.

\bibitem[Choi et~al.(2019)Choi, Xu, Li, Dusenberry, Flores, Xue, and
  Dai]{Choi2019}
E.~Choi, Z.~Xu, Y.~Li, M.~W. Dusenberry, G.~Flores, Y.~Xue, and A.~M. Dai.
\newblock {Graph Convolutional Transformer: Learning the Graphical Structure of
  Electronic Health Records}.
\newblock jun 2019.
\newblock URL \url{http://arxiv.org/abs/1906.04716}.

\bibitem[Choi et~al.(2016{\natexlab{d}})Choi, Chiu, and Sontag]{ChoiY2016}
Y.~Choi, C.~Y.-I. Chiu, and D.~Sontag.
\newblock {Learning Low-Dimensional Representations of Medical Concepts.}
\newblock \emph{AMIA Joint Summits on Translational Science proceedings. AMIA
  Joint Summits on Translational Science}, 2016:\penalty0 41--50,
  2016{\natexlab{d}}.
\newblock ISSN 2153-4063.
\newblock URL \url{http://www.ncbi.nlm.nih.gov/pubmed/27570647
  http://www.pubmedcentral.nih.gov/articlerender.fcgi?artid=PMC5001761}.

\bibitem[Chung and Woo(2018)]{Chung2018}
A.~S. Chung and H.~H. Woo.
\newblock {Update on minimally invasive surgery and benign prostatic
  hyperplasia}.
\newblock \emph{Asian Journal of Urology}, 5\penalty0 (1):\penalty0 22--27, jan
  2018.
\newblock ISSN 22143890.
\newblock \doi{10.1016/j.ajur.2017.06.001}.

\bibitem[Coorevits et~al.(2013)Coorevits, Sundgren, Klein, Bahr, Claerhout,
  Daniel, Dugas, Dupont, Schmidt, Singleton, De~Moor, and Kalra]{article}
P.~Coorevits, M.~Sundgren, G.~Klein, A.~Bahr, B.~Claerhout, C.~Daniel,
  M.~Dugas, D.~Dupont, A.~Schmidt, P.~Singleton, G.~De~Moor, and D.~Kalra.
\newblock Electronic health records: New opportunities for clinical research.
\newblock \emph{Journal of internal medicine}, 274, 08 2013.
\newblock \doi{10.1111/joim.12119}.

\bibitem[Cornu et~al.(2015)Cornu, Ahyai, Bachmann, {De La Rosette}, Gilling,
  Gratzke, McVary, Novara, Woo, and Madersbacher]{Cornu2015}
J.~N. Cornu, S.~Ahyai, A.~Bachmann, J.~{De La Rosette}, P.~Gilling, C.~Gratzke,
  K.~McVary, G.~Novara, H.~Woo, and S.~Madersbacher.
\newblock {A systematic review and meta-analysis of functional outcomes and
  complications following transurethral procedures for lower urinary tract
  symptoms resulting from benign prostatic obstruction: An update}.
\newblock \emph{European Urology}, 67\penalty0 (6):\penalty0 1066--1096, jun
  2015.
\newblock ISSN 18737560.
\newblock \doi{10.1016/j.eururo.2014.06.017}.

\bibitem[Cornu et~al.(2018)Cornu, Drake, Gacci, Gratzke, Herrmann,
  Madersbacher, Mamoulakis, and Tikkinen]{Cornu2018}
N.~Cornu, M.~Drake, M.~Gacci, C.~Gratzke, T.~Herrmann, S.~Madersbacher,
  C.~Mamoulakis, and K.~Tikkinen.
\newblock {EAU Guidelines: Management of Non-neurogenic Male LUTS | Uroweb}.
\newblock Technical report, 2018.
\newblock URL
  \url{https://uroweb.org/guideline/treatment-of-non-neurogenic-male-luts/}.

\bibitem[Dahm et~al.(2017)Dahm, Brasure, MacDonald, Olson, Nelson, Fink,
  Rwabasonga, Risk, and Wilt]{Dahm2017}
P.~Dahm, M.~Brasure, R.~MacDonald, C.~M. Olson, V.~A. Nelson, H.~A. Fink,
  B.~Rwabasonga, M.~C. Risk, and T.~J. Wilt.
\newblock {Comparative Effectiveness of Newer Medications for Lower Urinary
  Tract Symptoms Attributed to Benign Prostatic Hyperplasia: A Systematic
  Review and Meta-analysis}, apr 2017.
\newblock ISSN 18737560.

\bibitem[Dai et~al.(2019)Dai, Yang, Yang, Carbonell, Le, and
  Salakhutdinov]{dai_transformer-xl_2019}
Z.~Dai, Z.~Yang, Y.~Yang, J.~Carbonell, Q.~V. Le, and R.~Salakhutdinov.
\newblock Transformer-{XL}: {Attentive} {Language} {Models} {Beyond} a
  {Fixed}-{Length} {Context}.
\newblock \emph{arXiv:1901.02860 [cs, stat]}, June 2019.
\newblock URL \url{http://arxiv.org/abs/1901.02860}.
\newblock arXiv: 1901.02860.

\bibitem[Devlin et~al.(2018)Devlin, Chang, Lee, and Toutanova]{Devlin2018}
J.~Devlin, M.-W. Chang, K.~Lee, and K.~Toutanova.
\newblock {BERT: Pre-training of Deep Bidirectional Transformers for Language
  Understanding}.
\newblock \emph{North American Association for Computational Linguistics
  (NAACL)}, oct 2018.
\newblock URL \url{http://arxiv.org/abs/1810.04805}.

\bibitem[Diogo et~al.(2018)Diogo, Tian, Franklin, et~al.]{Diogo2018}
D.~Diogo, C.~Tian, C.~S. Franklin, et~al.
\newblock {Phenome-wide association studies across large population cohorts
  support drug target validation}.
\newblock \emph{Nature Communications}, 9\penalty0 (1), dec 2018.
\newblock ISSN 20411723.
\newblock \doi{10.1038/s41467-018-06540-3}.

\bibitem[Dozat(2016)]{Dozat2016}
T.~Dozat.
\newblock {Incorporating Nesterov Momentum into Adam}.
\newblock \emph{ICLR Workshop}, \penalty0 (1):\penalty0 2013--2016, 2016.

\bibitem[Fonteneau et~al.(2017)Fonteneau, {Le Meur}, Cohen-Akenine,
  et~al.]{Fonteneau2017}
L.~Fonteneau, N.~{Le Meur}, A.~Cohen-Akenine, et~al.
\newblock {The use of administrative health databases in infectious disease
  epidemiology and public health}.
\newblock \emph{Revue d'epidemiologie et de sante publique}, 65 Suppl
  4:\penalty0 S174--S182, oct 2017.
\newblock ISSN 0398-7620.
\newblock \doi{10.1016/j.respe.2017.03.131}.
\newblock URL \url{http://www.ncbi.nlm.nih.gov/pubmed/28624133}.

\bibitem[Hashim and Abrams(2015)]{Hashim2015}
H.~Hashim and P.~Abrams.
\newblock {Transurethral resection of the prostate for benign prostatic
  obstruction: Will it remain the gold standard?}
\newblock \emph{European Urology}, 67\penalty0 (6):\penalty0 1097--1098, jun
  2015.
\newblock ISSN 18737560.
\newblock \doi{10.1016/j.eururo.2014.12.022}.

\bibitem[Hochreiter and Schmidhuber(1997)]{hochreiter1997long}
S.~Hochreiter and J.~Schmidhuber.
\newblock Long short-term memory.
\newblock \emph{Neural computation}, 9\penalty0 (8):\penalty0 1735--1780, 1997.

\bibitem[Hub(2019)]{hdh}
H.~D. Hub.
\newblock \url{https://www.health-data-hub.fr}, 2019.

\bibitem[Kim et~al.(2016)Kim, Larson, and Andriole]{Kim2016}
E.~H. Kim, J.~A. Larson, and G.~L. Andriole.
\newblock {Management of Benign Prostatic Hyperplasia.}
\newblock \emph{Annual review of medicine}, 67:\penalty0 137--51, 2016.
\newblock ISSN 1545-326X.
\newblock \doi{10.1146/annurev-med-063014-123902}.
\newblock URL \url{http://www.ncbi.nlm.nih.gov/pubmed/26331999}.

\bibitem[Kingma and Ba(2015)]{Kingma2014}
D.~P. Kingma and J.~L. Ba.
\newblock {Adam: A method for stochastic optimization}.
\newblock \emph{3rd International Conference on Learning Representations, ICLR
  2015 - Conference Track Proceedings}, dec 2015.
\newblock URL \url{http://arxiv.org/abs/1412.6980}.

\bibitem[Kitaev et~al.(2020)Kitaev, Kaiser, and Levskaya]{kitaev_reformer_2020}
N.~Kitaev, L.~Kaiser, and A.~Levskaya.
\newblock Reformer: {The} {Efficient} {Transformer}.
\newblock \emph{arXiv:2001.04451 [cs, stat]}, Jan. 2020.
\newblock URL \url{http://arxiv.org/abs/2001.04451}.
\newblock arXiv: 2001.04451.

\bibitem[Lan et~al.(2020)Lan, Chen, Goodman, Gimpel, Sharma, and
  Soricut]{lan_albert_2020}
Z.~Lan, M.~Chen, S.~Goodman, K.~Gimpel, P.~Sharma, and R.~Soricut.
\newblock {ALBERT}: {A} {Lite} {BERT} for {Self}-supervised {Learning} of
  {Language} {Representations}.
\newblock \emph{arXiv:1909.11942 [cs]}, Jan. 2020.
\newblock URL \url{http://arxiv.org/abs/1909.11942}.
\newblock arXiv: 1909.11942.

\bibitem[LeCun et~al.(2012)LeCun, Bottou, Orr, and
  Müller]{lecun_efficient_2012}
Y.~A. LeCun, L.~Bottou, G.~B. Orr, and K.-R. Müller.
\newblock Efficient {BackProp}.
\newblock In G.~Montavon, G.~B. Orr, and K.-R. Müller, editors, \emph{Neural
  {Networks}: {Tricks} of the {Trade}: {Second} {Edition}}, pages 9--48.
  Springer Berlin Heidelberg, Berlin, Heidelberg, 2012.
\newblock ISBN 978-3-642-35289-8.
\newblock \doi{10.1007/978-3-642-35289-8_3}.
\newblock URL \url{https://doi.org/10.1007/978-3-642-35289-8_3}.

\bibitem[Li et~al.(2019)Li, Rao, Solares, Hassaine, Canoy, Zhu, Rahimi, and
  Salimi-Khorshidi]{Li2019}
Y.~Li, S.~Rao, J.~R.~A. Solares, A.~Hassaine, D.~Canoy, Y.~Zhu, K.~Rahimi, and
  G.~Salimi-Khorshidi.
\newblock {BEHRT: Transformer for Electronic Health Records}.
\newblock jul 2019.
\newblock URL \url{http://arxiv.org/abs/1907.09538}.

\bibitem[Lin et~al.(2015)Lin, Kuo, Hung, Wu, Chen, Yu, Hsu, Lee, Chen, and
  Hwang]{lin_association_2015}
M.~Y. Lin, M.~C. Kuo, C.~C. Hung, W.~J. Wu, L.~T. Chen, M.~L. Yu, C.-C. Hsu,
  C.-H. Lee, H.-C. Chen, and S.-J. Hwang.
\newblock Association of {Dialysis} with the {Risks} of {Cancers}.
\newblock \emph{PLoS ONE}, 10\penalty0 (4), Apr. 2015.
\newblock ISSN 1932-6203.
\newblock \doi{10.1371/journal.pone.0122856}.
\newblock URL \url{https://www.ncbi.nlm.nih.gov/pmc/articles/PMC4395337/}.

\bibitem[Lin et~al.(2019)Lin, Feng, {Dos Santos}, Yu, Xiang, Zhou, and
  Bengio]{Lin2019}
Z.~Lin, M.~Feng, C.~N. {Dos Santos}, M.~Yu, B.~Xiang, B.~Zhou, and Y.~Bengio.
\newblock {A structured self-attentive sentence embedding}.
\newblock \emph{5th International Conference on Learning Representations, ICLR
  2017 - Conference Track Proceedings}, 2019.
\newblock URL \url{http://arxiv.org/abs/1703.03130}.

\bibitem[Lipton et~al.(2016)Lipton, Kale, Elkan, and Wetzel]{Lipton2015}
Z.~C. Lipton, D.~C. Kale, C.~Elkan, and R.~Wetzel.
\newblock {Learning to diagnose with LSTM recurrent neural networks}.
\newblock \emph{4th International Conference on Learning Representations, ICLR
  2016 - Conference Track Proceedings}, nov 2016.
\newblock URL \url{http://arxiv.org/abs/1511.03677}.

\bibitem[Lourenco et~al.(2010)Lourenco, Shaw, Fraser, MacLennan, N'Dow, and
  Pickard]{Lourenco2010}
T.~Lourenco, M.~Shaw, C.~Fraser, G.~MacLennan, J.~N'Dow, and R.~Pickard.
\newblock {The clinical effectiveness of transurethral incision of the
  prostate: a systematic review of randomised controlled trials.}
\newblock \emph{World journal of urology}, 28\penalty0 (1):\penalty0 23--32,
  feb 2010.
\newblock ISSN 1433-8726.
\newblock \doi{10.1007/s00345-009-0496-8}.
\newblock URL \url{http://www.ncbi.nlm.nih.gov/pubmed/20033744}.

\bibitem[Luo et~al.(2019)Luo, Xu, and Carin]{Luo2019}
D.~Luo, H.~Xu, and L.~Carin.
\newblock {Interpretable ICD Code Embeddings with Self- and Mutual-Attention
  Mechanisms}.
\newblock jun 2019.
\newblock URL \url{http://arxiv.org/abs/1906.05492}.

\bibitem[Ma et~al.(2017)Ma, Chitta, Zhou, You, Sun, and Gao]{Ma2017}
F.~Ma, R.~Chitta, J.~Zhou, Q.~You, T.~Sun, and J.~Gao.
\newblock {Dipole: Diagnosis prediction in healthcare via attention-based
  bidirectional recurrent neural networks}.
\newblock In \emph{Proceedings of the ACM SIGKDD International Conference on
  Knowledge Discovery and Data Mining}, volume Part F129685, pages 1903--1911.
  Association for Computing Machinery, aug 2017.
\newblock ISBN 9781450348874.
\newblock \doi{10.1145/3097983.3098088}.

\bibitem[Macey and Raynor(2016)]{Macey2016}
M.~R. Macey and M.~C. Raynor.
\newblock {Medical and Surgical Treatment Modalities for Lower Urinary Tract
  Symptoms in the Male Patient Secondary to Benign Prostatic Hyperplasia: A
  Review}.
\newblock \emph{Seminars in Interventional Radiology}, 33\penalty0
  (3):\penalty0 217--223, sep 2016.
\newblock ISSN 10988963.
\newblock \doi{10.1055/s-0036-1586142}.

\bibitem[Madersbacher et~al.(2005)Madersbacher, Lackner, Br{\"{o}}ssner,
  R{\"{o}}hlich, Stancik, Willinger, Schatzl, and {Prostate Study Group of the
  Austrian Society of Urology}]{Madersbacher2005}
S.~Madersbacher, J.~Lackner, C.~Br{\"{o}}ssner, M.~R{\"{o}}hlich, I.~Stancik,
  M.~Willinger, G.~Schatzl, and {Prostate Study Group of the Austrian Society
  of Urology}.
\newblock {Reoperation, myocardial infarction and mortality after transurethral
  and open prostatectomy: a nation-wide, long-term analysis of 23,123 cases.}
\newblock \emph{European urology}, 47\penalty0 (4):\penalty0 499--504, apr
  2005.
\newblock ISSN 0302-2838.
\newblock \doi{10.1016/j.eururo.2004.12.010}.
\newblock URL \url{http://www.ncbi.nlm.nih.gov/pubmed/15774249}.

\bibitem[Mallik and Zhao(2020)]{mallik_graph_2020}
S.~Mallik and Z.~Zhao.
\newblock Graph- and rule-based learning algorithms: A comprehensive review of
  their applications for cancer type classification and prognosis using genomic
  data.
\newblock \emph{Briefings in Bioinformatics}, 21\penalty0 (2):\penalty0
  368--394, Mar. 2020.
\newblock ISSN 1467-5463, 1477-4054.
\newblock \doi{10.1093/bib/bby120}.

\bibitem[McInnes et~al.(2018)McInnes, Healy, and Melville]{mcinnes_umap_2018}
L.~McInnes, J.~Healy, and J.~Melville.
\newblock {UMAP}: {Uniform} {Manifold} {Approximation} and {Projection} for
  {Dimension} {Reduction}.
\newblock \emph{arXiv:1802.03426 [cs, stat]}, Dec. 2018.
\newblock URL \url{http://arxiv.org/abs/1802.03426}.
\newblock arXiv: 1802.03426.

\bibitem[Michel et~al.(2004)Michel, Heemann, Schumacher, Mehlburger, and
  Goepel]{michel_association_2004}
M.~C. Michel, U.~Heemann, H.~Schumacher, L.~Mehlburger, and M.~Goepel.
\newblock Association of hypertension with symptoms of benign prostatic
  hyperplasia.
\newblock \emph{The Journal of Urology}, 172\penalty0 (4 Pt 1):\penalty0
  1390--1393, Oct. 2004.
\newblock ISSN 0022-5347.
\newblock \doi{10.1097/01.ju.0000139995.85780.d8}.

\bibitem[Mikolov et~al.(2013)Mikolov, Sutskever, Chen, Corrado, and
  Dean]{mikolov2013distributed}
T.~Mikolov, I.~Sutskever, K.~Chen, G.~S. Corrado, and J.~Dean.
\newblock Distributed representations of words and phrases and their
  compositionality.
\newblock In \emph{Advances in neural information processing systems}, pages
  3111--3119, 2013.

\bibitem[Miotto et~al.(2016)Miotto, Li, Kidd, and Dudley]{Miotto2016}
R.~Miotto, L.~Li, B.~A. Kidd, and J.~T. Dudley.
\newblock {Deep Patient: An Unsupervised Representation to Predict the Future
  of Patients from the Electronic Health Records}.
\newblock \emph{Scientific Reports}, 6, may 2016.
\newblock ISSN 20452322.
\newblock \doi{10.1038/srep26094}.

\bibitem[Morel et~al.(2019)Morel, Bacry, Ga{\"{i}}ffas, Guilloux, and
  Leroy]{Morel2019}
M.~Morel, E.~Bacry, S.~Ga{\"{i}}ffas, A.~Guilloux, and F.~Leroy.
\newblock {ConvSCCS: convolutional self-controlled case series model for lagged
  adverse event detection.}
\newblock \emph{Biostatistics (Oxford, England)}, mar 2019.
\newblock ISSN 1468-4357.
\newblock \doi{10.1093/biostatistics/kxz003}.
\newblock URL \url{http://www.ncbi.nlm.nih.gov/pubmed/30851046}.

\bibitem[Neil et~al.(2016)Neil, Pfeiffer, and Liu]{Neil2016}
D.~Neil, M.~Pfeiffer, and S.~C. Liu.
\newblock {Phased LSTM: Accelerating recurrent network training for long or
  event-based sequences}.
\newblock \emph{Advances in Neural Information Processing Systems}, pages
  3889--3897, oct 2016.
\newblock ISSN 10495258.
\newblock URL \url{http://arxiv.org/abs/1610.09513}.

\bibitem[Nguyen et~al.(2017)Nguyen, Tran, Wickramasinghe, and
  Venkatesh]{Nguyen2016}
P.~Nguyen, T.~Tran, N.~Wickramasinghe, and S.~Venkatesh.
\newblock {Deepr: A Convolutional Net for Medical Records}.
\newblock \emph{IEEE Journal of Biomedical and Health Informatics}, 21\penalty0
  (1):\penalty0 22--30, jul 2017.
\newblock ISSN 21682194.
\newblock \doi{10.1109/JBHI.2016.2633963}.
\newblock URL \url{http://arxiv.org/abs/1607.07519}.

\bibitem[Pedregosa et~al.(2011)Pedregosa, Varoquaux, Gramfort, Michel, Thirion,
  Grisel, Blondel, Prettenhofer, Weiss, Dubourg, et~al.]{pedregosa2011scikit}
F.~Pedregosa, G.~Varoquaux, A.~Gramfort, V.~Michel, B.~Thirion, O.~Grisel,
  M.~Blondel, P.~Prettenhofer, R.~Weiss, V.~Dubourg, et~al.
\newblock Scikit-learn: Machine learning in python.
\newblock \emph{Journal of machine learning research}, 12\penalty0
  (Oct):\penalty0 2825--2830, 2011.

\bibitem[Pennington et~al.(2014)Pennington, Socher, and
  Manning]{pennington2014}
J.~Pennington, R.~Socher, and C.~Manning.
\newblock {G}love: Global vectors for word representation.
\newblock In \emph{Proceedings of the 2014 Conference on Empirical Methods in
  Natural Language Processing ({EMNLP})}, pages 1532--1543, Doha, Qatar, Oct.
  2014. Association for Computational Linguistics.
\newblock \doi{10.3115/v1/D14-1162}.
\newblock URL \url{https://www.aclweb.org/anthology/D14-1162}.

\bibitem[Pham et~al.(2017)Pham, Tran, Phung, and Venkatesh]{Pham2017}
T.~Pham, T.~Tran, D.~Phung, and S.~Venkatesh.
\newblock {Predicting healthcare trajectories from medical records: A deep
  learning approach.}
\newblock \emph{Journal of biomedical informatics}, 69:\penalty0 218--229,
  2017.
\newblock ISSN 1532-0480.
\newblock \doi{10.1016/j.jbi.2017.04.001}.
\newblock URL \url{http://www.ncbi.nlm.nih.gov/pubmed/28410981}.

\bibitem[Rajkomar et~al.(2018)Rajkomar, Oren, Chen, et~al.]{Rajkomar2018}
A.~Rajkomar, E.~Oren, K.~Chen, et~al.
\newblock {Scalable and accurate deep learning with electronic health records}.
\newblock \emph{npj Digital Medicine}, 1\penalty0 (1), jan 2018.
\newblock ISSN 2398-6352.
\newblock \doi{10.1038/s41746-018-0029-1}.
\newblock URL \url{https://arxiv.org/abs/1801.07860}.

\bibitem[Scailteux et~al.(2019)Scailteux, Droitcourt, Balusson, Nowak, Kerbrat,
  Dupuy, Drezen, Happe, and Oger]{Scailteux2019}
L.-M. Scailteux, C.~Droitcourt, F.~Balusson, E.~Nowak, S.~Kerbrat, A.~Dupuy,
  E.~Drezen, A.~Happe, and E.~Oger.
\newblock {French administrative health care database (SNDS): The value of its
  enrichment.}
\newblock \emph{Therapie}, 74\penalty0 (2):\penalty0 215--223, apr 2019.
\newblock ISSN 1958-5578.
\newblock \doi{10.1016/j.therap.2018.09.072}.
\newblock URL \url{http://www.ncbi.nlm.nih.gov/pubmed/30392702}.

\bibitem[Schuster and Paliwal(1997)]{schuster1997bidirectional}
M.~Schuster and K.~K. Paliwal.
\newblock Bidirectional recurrent neural networks.
\newblock \emph{IEEE transactions on Signal Processing}, 45\penalty0
  (11):\penalty0 2673--2681, 1997.

\bibitem[Shickel et~al.(2018)Shickel, Tighe, Bihorac, and Rashidi]{Shickel2018}
B.~Shickel, P.~J. Tighe, A.~Bihorac, and P.~Rashidi.
\newblock {Deep EHR: A Survey of Recent Advances in Deep Learning Techniques
  for Electronic Health Record (EHR) Analysis}.
\newblock \emph{IEEE Journal of Biomedical and Health Informatics}, 22\penalty0
  (5):\penalty0 1589--1604, sep 2018.
\newblock ISSN 21682194.
\newblock \doi{10.1109/JBHI.2017.2767063}.

\bibitem[Silva et~al.(2014)Silva, Silva, and Cruz]{Silva2014}
J.~Silva, C.~M. Silva, and F.~Cruz.
\newblock {Current medical treatment of lower urinary tract symptoms/BPH: do we
  have a standard?}
\newblock \emph{Current opinion in urology}, 24\penalty0 (1):\penalty0 21--8,
  jan 2014.
\newblock ISSN 1473-6586.
\newblock \doi{10.1097/MOU.0000000000000007}.
\newblock URL \url{http://www.ncbi.nlm.nih.gov/pubmed/24231531}.

\bibitem[SNDS(2019)]{SNDS2019}
SNDS.
\newblock {Processus d'acc{\`{e}}s aux donn{\'{e}}es | SNDS}, 2019.
\newblock URL
  \url{https://www.snds.gouv.fr/SNDS/Processus-d-acces-aux-donnees}.

\bibitem[Taborelli et~al.(2019)Taborelli, Toffolutti, Del~Zotto,
  et~al.]{taborelli_increased_2019}
M.~Taborelli, F.~Toffolutti, S.~Del~Zotto, et~al.
\newblock Increased cancer risk in patients undergoing dialysis: a
  population-based cohort study in {North}-{Eastern} {Italy}.
\newblock \emph{BMC Nephrology}, 20\penalty0 (1):\penalty0 107, Mar. 2019.
\newblock ISSN 1471-2369.
\newblock \doi{10.1186/s12882-019-1283-4}.
\newblock URL \url{https://doi.org/10.1186/s12882-019-1283-4}.

\bibitem[Tuppin et~al.(2017)Tuppin, Rudant, Constantinou,
  Gastaldi-M{\'{e}}nager, Rachas, de~Roquefeuil, Maura, Caillol, Tajahmady,
  Coste, Gissot, Weill, and Fagot-Campagna]{Tuppin2017}
P.~Tuppin, J.~Rudant, P.~Constantinou, C.~Gastaldi-M{\'{e}}nager, A.~Rachas,
  L.~de~Roquefeuil, G.~Maura, H.~Caillol, A.~Tajahmady, J.~Coste, C.~Gissot,
  A.~Weill, and A.~Fagot-Campagna.
\newblock {Value of a national administrative database to guide public
  decisions: From the syst{\`{e}}me national d'information interr{\'{e}}gimes
  de l'Assurance Maladie (SNIIRAM) to the syst{\`{e}}me national des
  donn{\'{e}}es de sant{\'{e}} (SNDS) in France.}
\newblock \emph{Revue d'epidemiologie et de sante publique}, 65 Suppl
  4:\penalty0 S149--S167, oct 2017.
\newblock ISSN 0398-7620.
\newblock \doi{10.1016/j.respe.2017.05.004}.
\newblock URL \url{http://www.ncbi.nlm.nih.gov/pubmed/28756037}.

\bibitem[Vaswani et~al.(2017)Vaswani, Shazeer, Parmar, Uszkoreit, Jones, Gomez,
  Kaiser, and Polosukhin]{Vaswani2017}
A.~Vaswani, N.~Shazeer, N.~Parmar, J.~Uszkoreit, L.~Jones, A.~N. Gomez,
  {\L}.~Kaiser, and I.~Polosukhin.
\newblock {Attention is all you need}.
\newblock \emph{Advances in Neural Information Processing Systems},
  2017-December:\penalty0 5999--6009, 2017.
\newblock ISSN 10495258.

\bibitem[Wan et~al.(2013)Wan, Zeiler, Zhang, Cun, and Fergus]{Wan2013}
L.~Wan, M.~Zeiler, S.~Zhang, Y.~L. Cun, and R.~Fergus.
\newblock Regularization of neural networks using dropconnect.
\newblock In S.~Dasgupta and D.~McAllester, editors, \emph{Proceedings of the
  30th International Conference on Machine Learning}, volume~28 of
  \emph{Proceedings of Machine Learning Research}, pages 1058--1066, Atlanta,
  Georgia, USA, Jun 2013. PMLR.
\newblock URL \url{http://proceedings.mlr.press/v28/wan13.html}.

\bibitem[Wu et~al.(2019)Wu, Xiong, Schnabel, Zhang, Wang, and
  Bennett]{wu_combiner_2019}
J.~Wu, C.~Xiong, T.~Schnabel, Y.~Zhang, W.~Y. Wang, and P.~Bennett.
\newblock Combiner: {Inductively} {Learning} {Tree} {Structured} {Attention} in
  {Transformers}.
\newblock Sept. 2019.
\newblock URL \url{https://openreview.net/forum?id=B1eySTVtvB}.

\bibitem[Xiao et~al.(2018)Xiao, Choi, and Sun]{Xiao2018}
C.~Xiao, E.~Choi, and J.~Sun.
\newblock {Opportunities and challenges in developing deep learning models
  using electronic health records data: a systematic review.}
\newblock \emph{Journal of the American Medical Informatics Association :
  JAMIA}, 25\penalty0 (10):\penalty0 1419--1428, oct 2018.
\newblock ISSN 1527-974X.
\newblock \doi{10.1093/jamia/ocy068}.
\newblock URL \url{http://www.ncbi.nlm.nih.gov/pubmed/29893864
  http://www.pubmedcentral.nih.gov/articlerender.fcgi?artid=PMC6188527}.

\bibitem[Zhang et~al.(2018)Zhang, Kowsari, Harrison, Lobo, and
  Barnes]{Zhang2018}
J.~Zhang, K.~Kowsari, J.~H. Harrison, J.~M. Lobo, and L.~E. Barnes.
\newblock {Patient2Vec: A Personalized Interpretable Deep Representation of the
  Longitudinal Electronic Health Record}.
\newblock \emph{IEEE Access}, 6:\penalty0 65333--65346, 2018.
\newblock ISSN 21693536.
\newblock \doi{10.1109/ACCESS.2018.2875677}.

\bibitem[Zhu et~al.(2017)Zhu, Li, Liao, Wang, Guan, Liu, and Cai]{Zhu2017time}
Y.~Zhu, H.~Li, Y.~Liao, B.~Wang, Z.~Guan, H.~Liu, and D.~Cai.
\newblock What to do next: Modeling user behaviors by time-lstm.
\newblock In \emph{Proceedings of the Twenty-Sixth International Joint
  Conference on Artificial Intelligence, {IJCAI-17}}, pages 3602--3608, 2017.
\newblock \doi{10.24963/ijcai.2017/504}.
\newblock URL \url{https://doi.org/10.24963/ijcai.2017/504}.

\end{thebibliography}
\end{document}